\documentclass[11pt]{article}

\usepackage[margin=1in]{geometry}
\usepackage{lipsum}

\usepackage[utf8]{inputenc} % allow utf-8 input
\usepackage[T1]{fontenc}    % use 8-bit T1 fonts
\usepackage{tgtermes}

\usepackage{url}            % simple URL typesetting
\usepackage{booktabs}       % professional-quality tables
\usepackage{amsfonts}       % blackboard math symbols
\usepackage{nicefrac}       % compact symbols for 1/2, etc.
\usepackage{microtype}      % microtypography

\usepackage[english]{babel}
\usepackage{subcaption}

\usepackage{amsmath}
\usepackage{enumitem}
\usepackage{amssymb}
\usepackage{graphicx}
\usepackage{bm}
\usepackage{theorem}
\usepackage[colorlinks=true, allcolors=blue]{hyperref}

\usepackage{mathtools}
\newenvironment{proof}{\paragraph{\it Proof.}}{\hfill$\square$}

\usepackage[title]{appendix}
\usepackage{comment}
\usepackage{amsfonts}
\usepackage{dsfont}
\usepackage{hyperref}
\usepackage{lipsum}
\usepackage{dsfont,subcaption,float}
\usepackage{algorithm}
\usepackage{algorithmic}

\usepackage{thm-restate}
\usepackage[size=small,labelfont=bf]{caption}

\usepackage[table]{xcolor}
\theoremstyle{plain}
\newtheorem{theorem}{Theorem}[section]

\newtheorem{lemma}[theorem]{Lemma}
\newtheorem{corollary}[theorem]{Corollary}
\theoremstyle{definition}
\newtheorem{definition}[theorem]{Definition}
\newtheorem{assumption}{Assumption}
\theoremstyle{remark}

\newcommand{\TV}{d_{\texttt{TV}}}

\newcommand{\poly}{\mathrm{poly}}

\newcommand{\Ber}{\mathrm{Ber}}

\newcommand{\HL}{d_\texttt{H}}
\newcommand{\KL}{\textrm{KL}} 

\newcommand{\mC}{\mathcal{C}}	% Confidence
\newcommand{\mS}{\mathcal{S}}	% States
\newcommand{\mA}{\mathcal{A}}	% Actions
\newcommand{\mO}{\mathcal{O}}	% Observations
\newcommand{\mD}{\mathcal{D}}	% Data 
\newcommand{\mM}{\mathcal{M}}	% Model class
\newcommand{\mG}{\mathcal{G}}	% Generalized linear function class
\newcommand{\mI}{\mathcal{I}}	% real space   
\newcommand{\indic}[1]{\mathds{1}\left\{ #1 \right\}} % indicator function 
\newcommand{\diag}{\textbf{diag}}

\newcommand{\PP}{\mathds{P}}    % probability 

\newcommand{\Exs}{\mathbb{E}}

\newcommand{\sign}{\texttt{sgn}}

\newcommand{\tssum}{\textstyle \sum}

\allowdisplaybreaks

\newif\ifdraft
\drafttrue  % comment this line to enabble draft mode

\newif\ifarxiv

\arxivtrue
\usepackage{natbib}

\title{\bf{\LARGE{Prospective Side Information for Latent MDPs}}}

\usepackage{authblk}
\author[1]{Jeongyeol Kwon}
\author[2]{Yonathan Efroni}
\author[3]{Shie Mannor}
\author[4]{Constantine Caramanis}
\affil[1]{Wisconsin Institute for Discovery, UW-Madison}
\affil[2]{Meta, New York}
\affil[3]{Department of Electrical Engineering, Technion / NVIDIA}
\affil[4]{Department of Electrical and Computer Engineering, UT Austin}

\begin{document}

\maketitle

\begin{abstract}
In many interactive decision-making settings, there is latent and unobserved information that remains fixed. Consider, for example, a dialogue system, where complete information about a user, such as the user's preferences, is not given. In such an environment, the latent information remains fixed throughout each episode, since the identity of the user does not change during an interaction. This type of environment can be modeled as a Latent Markov Decision Process (LMDP), a special instance of Partially Observed Markov Decision Processes (POMDPs). Recently,~\citet{kwon2021rl} established exponential lower bounds in the number of latent contexts for the LMDP class. This puts forward a question: under which natural assumptions a near-optimal policy of an LMDP can be efficiently learned? In this work, we study the class of LMDPs with  {\em prospective side information}, when an agent receives additional, weakly revealing, information on the latent context at the beginning of each episode. We show that, surprisingly, this problem is not captured by contemporary settings and algorithms designed for partially observed environments. We then establish that any sample efficient algorithm must suffer at least $\Omega(K^{2/3})$-regret, as opposed to standard $\Omega(\sqrt{K})$ lower bounds, and design an algorithm with a matching upper bound.

\end{abstract}

\section{Introduction}

% \jycomment{Just wrote an idea per paragraph}

% \gencomment{- General RL \cite{sutton2018reinforcement} with Partial observation \cite{smallwood1973optimal, krishnamurthy2016pac} is hard.}

Many real-world sequential decision problems are partially observed, and full information on the state of the system is not known. In its full generality, such a setting can be formulated as a Partially Observed Markov Decision Process (POMDP). Due to its prevalence, POMDPs have been extensively studied in past decades \cite{smallwood1973optimal, pineau2006anytime}. Yet, with no further assumptions, it is known to be hard from both computational and learnability perspectives \cite{papadimitriou1987complexity, krishnamurthy2016pac}. A possible meaningful way moving forward is to restrict the study to special and widespread classes of POMDPs. Recent advances in literature put efforts into identifying tractable subclasses of POMDPs in several aspects \cite{dann2018oracle, du2019provably, efroni2022provable, uehara2022provably, kwon2021rl, liu2022partially}.

% \gencomment{- We consider LMDP framework \cite{kwon2021rl} since this is well-motivated model in both theory and practice. - Even learning in this subclass of POMDPs is challenging. In particular, without any assumptions, the worst-case lower bound is $(SA)^{\Omega(M)}$. This hardness result is disappointing, especially when the super-polynomial bound $(SA)^{\omega(1)}$ exists even when the transition dynamics are shared in all environments \cite{kwon2023reward}. }

We consider a partially observed sequential problem in which the latent information remains fixed during each episodic interaction  \cite{chades2012momdps, hallak2015contextual, brunskill2013sample, steimle2018multi,kwon2021rl}, also referred as Latent MDPs (LMDPs). Such a setting can model many common problems, e.g., dialogue and recommender systems, when complete information on a user is not given, yet, each user remains fixed within each episodic interaction. Recently, \citet{kwon2021rl, kwon2023reward} derived exponential worst-case lower bounds in the number of contexts for this subclass of POMDPs. This implies that, in general, near optimal policy of an LMDP cannot be learned efficiently when the number of latent context is large. 

% \gencomment{- It is important to consider additional but practical assumptions to break the super-polynomial barrier. Fortunately, there are many cases that we can start with initial side information about the environment. Gender, Age, Symptoms, Backgrounds, etc... - The goal of this paper is to show that in fact, there are (significant) benefits of side-information given at the initial time. We show that if side information is weakly revealing, then we can break the super-polynomial barrier in general.}

Under which assumptions, common in practice, do LMDPs can be efficiently learned? Prior work~\cite{kwon2021rl,zhou2022horizon, lee2023learning} established that given complete information on the latent context in hindsight, {\it i.e.}, at the end of each episode, LMDPs can be learned efficiently. In this work, we study a somewhat dual assumption; we assume that an agent can observe a weakly revealing side information on the latent context at the beginning of each episode, and refer it as {\it LMDP with  Prospective Side Information, or as LMDP-$\Psi$}. Differently than the case information is available in hindsight, for the LMDP-$\Psi$ setting, the policy can utilize the additional hint within each episode. We study lower bounds and matching upper bounds for this class of problems, and show this setting is tractable from the sample complexity perspective.

{\ifarxiv % ARXIV VERSION
    \begin{table}[t]
        \centering
        \begin{tabular}{ |c|c|c|>{\columncolor{green!20}}c|c|c| } 
            \hline
            \ & MDP  & $\alpha$-Revealing POMDP  & LMDP with $\alpha$-Prospective SI & LMDP & POMDP \\
            \hline
            UB & $\sqrt{A K}$ & $\poly(A,\alpha^{-1}) \sqrt{K}$ &  {\bf $\poly(A, \alpha^{-1}) K^{2/3}$}& --- & $A^H \sqrt{K}$  \\ 
            \hline 
            LB & $\sqrt{A K}$  & $\poly(A, \alpha^{-1}) \sqrt{K}$  & {\bf $\min\left(A^{\Omega(M)} \sqrt{K}, \poly(A, \alpha^{-1}) K^{2/3}\right)$} & $A^{M} \sqrt{K}$ & $A^{H} \sqrt{K}$ \\ 
            \hline
        \end{tabular}
        \caption{Regret upper and lower bounds in different classes of POMDPs, ordered by their degree of difficulty from the simplest to hardest (left to right). Dependencies on other problem parameters are omitted ({\it e.g.} $S$ and $H$). The results and the setting introduced in this work are highlighted in green.}
        % Complexities of general MDPs and POMDPs are well-known (see {\it e.g.,} \cite{jaksch2010near, krishnamurthy2016pac}). 
        % For $\alpha$-revealing POMDPs \cite{liu2022partially,chen2022partially}, $\sqrt{K}$ regret can be attained. Without any assumptions, the worst case regret of the LMDP class is  exponential in the number of latent contexts. No upper bound is currently known for general LMDPs.}
        \label{table:regret_result}
    \end{table}
\else % CONFERENCE VERSION
\begin{table*}[t]
    \centering
    \begin{tabular}{ |c|c|c|>{\columncolor{green!20}}c|c|c| } 
        \hline
        \ & MDP  & $\alpha$-Revealing POMDP  & LMDP with $\alpha$-Prospective SI & LMDP & POMDP \\
        \hline
        Upper Bound & $\sqrt{A K}$ & $\poly(A,\alpha^{-1}) \sqrt{K}$ &  {\bf $\poly(A, \alpha^{-1}) K^{2/3}$}& Unknown & $A^H \sqrt{K}$  \\ 
        \hline 
        Lower Bound & $\sqrt{A K}$  & $\poly(A, \alpha^{-1}) \sqrt{K}$  & {\bf $\min\left(A^{\Omega(M)} \sqrt{K}, \poly(A, \alpha^{-1}) K^{2/3}\right)$} & $A^{\Omega(M)} \sqrt{K}$ & $A^{\Omega(H)} \sqrt{K}$ \\ 
        \hline
    \end{tabular}
    \caption{Known regret upper and lower bounds in different classes of POMDPs, ordered by their degree of difficulty from the simplest to hardest (left to right). Dependencies on other problem parameters are omitted ({\it e.g.} $S$ and $H$). The results and the setting introduced in this work are highlighted in green.}
    % Complexities of general MDPs and POMDPs are well-known (see {\it e.g.,} \cite{jaksch2010near, krishnamurthy2016pac}). 
    % For $\alpha$-revealing POMDPs \cite{liu2022partially,chen2022partially}, $\sqrt{K}$ regret can be attained. Without any assumptions, the worst case regret of the LMDP class is  exponential in the number of latent contexts. No upper bound is currently known for general LMDPs.}
    \label{table:regret_result}
\end{table*}
\fi}

Technically speaking, our work builds upon recent algorithmic advancements for POMDPs \cite{liu2023optimistic, uehara2022provably, huang2023provably}. However, proper application of these requires care. As we show, the {LMDP-$\Psi$} class is not contained within the class of POMDPs previously known to be efficiently learnable, and, in fact, goes beyond common POMDP modeling assumptions. 
% This effect also differentiates the problem from the well-conditioned PSR, since the weakly-revealing information about the latent state cannot be used during the exploration phase. 
Further, we highlight a surprising limitation of exploiting the prospective side information on the latent context.
% However, even without the necessity to perform a test policy to obtain sufficient statistics of latent states (the latent context in our case),

\paragraph{Our Contributions} The main contributions of this work are the following (see also Table \ref{table:regret_result}). We introduce the problem of learning a near-optimal policy with $\alpha$-prospective side information for LMDPs, when the prospective side information weakly reveals information on the true latent state. We provide  a  $\poly(A, \alpha^{-1}) K^{2/3}$ regret upper bound, by building upon the pure exploration scheme developed in \citet{huang2023provably}. Namely, our upper bound does not suffer exponential dependence in the number of latent contexts. We also provide a lower bound of $\Omega\left(\frac{A}{\alpha^2 \epsilon^2} K^{2/3}\right)$ to this problem, unlike the $K^{1/2}$ rate one may expect. 
%\begin{itemize}
    %\item While a similar lower bound of $K^{2/3}$ regret has been reported for learning a multi-step weakly observable POMDPs~\cite{chen2023lower}. %, our  result shows that we can still suffer from $K^{2/3}$ regret lower bound even when the weakly-revealing information is given ``for free''.\yecomment{this sounds a bit vague} 
%\end{itemize}

\section{Preliminaries}
An episodic LMDP is defined as follows:
\begin{definition}[Latent MDP]
    \label{definition:lmdp_so}
    % Consider a family of LMDP instances defined on $(\mS, \mA, \mO)$, where $\mS$ is a set of states, $\mA$ is a set of actions, $\mO$ is a set of observations.
 An LMDP instance 
 % $\mM$
 consists of a tuple $\theta := \left(\{p_m\}_{m=1}^M, \{\mathbb{T}_m\}_{m=1}^M, \{\mathbb{O}_m\}_{m=1}^M \right)$, where $M$ is the number of latent contexts; $\{p_m\}_{m=1}^M$ are the mixing weights, the probability latent context $m$ is drawn at the beginning of an episode;  $\mathbb{T}_m \in \mathbb{R}^{S \times S\times A}, \mathbb{O}_m \in \mathbb{R}^{|\mO| \times S\times A}$ are the transition probabilities and instant observation distribution of $m^{th}$ MDPs, {\it i.e.,} $\mathbb{T}_m(s'|s,a) := \PP(s'|m, s,a)$ and $\mathbb{O}_m(o, s, a) := \PP(o|m,s,a)$ for state $s \in \mS$, next state $s'\in \mS$, action $a \in \mA$, instanteneous observation $o \in \mO$, and latent context $m\in [M]$.
\end{definition}
% We do not assume prior knowledge of the side-observation matrix, mixing weights, or any underlying model parameters. 
We assume that for all $o \in \mathcal{O}$, there is a known reward-decoding function $r: \mO \rightarrow \mathbb{R}$, and each reward is bounded $|r(o)| \le 1$. To simplify the discussion, we assume that the set of LMDP instances $\Theta$ has finite (but exponentially large) cardinality $|\Theta|$. Similarly, we also assume that the observation space is discrete and finite:
\begin{assumption}[Observation Space]
\label{assumption:reward_dist}
    Each observation attains a value in the set $\mathcal{O}$ which has finite but could be arbitrarily large cardinality $|\mO|$. 
\end{assumption}
All claims made in this paper hold similarly for the continuous model class with a standard $\epsilon$-discretization of $\Theta$ with the extra discretization error analysis similar to~\citet{liu2022partially} and continuous observations~\citet{liu2023optimistic}. 

 At the beginning of every episode, a latent and unobserved context $m \in [M]$ is sampled from a mixing distribution $\{p_m\}_{m=1}^M$ and is fixed for $H$ time steps. Without loss of generality, we assume that the system starts from time-step $t=0$ at a fixed initial state $s_{\text{dummy}}$ and transits to other states following the initial state distribution of the chosen MDP regardless of taken actions (and we always see a dummy observation $o_{\text{dummy}}$).

\paragraph{Prospective Side Information for LMDPs.}
In this work, we assume the LMDP is augmented with prospective side information.
Prospective side information is an additional observation given {\it prior to the beginning of the episode} and remains fixed along a trajectory.  Let $\mathcal{I}$ be the set of prospective side information values, and is assumed to be finite but may be arbitrarily large. Let $\mathbb{I} \in \mathbb{R}^{|\mI| \times M}$ be a context dependent emission matrix, {\it i.e.,} $\mathbb{I} (\iota,m) := \PP(\iota|m)$. We assume that the prospective side information is given only at the beginning of each episode and remains fixed during the entire episode. Further, we assume it provides some hint on the identity of the true latent MDP. Formally, we assume the following weakly revealing condition:
\begin{assumption}[Prospective Side Information]
    \label{assumption:full_rank_obs}
    For any two belief vectors $\bar{v}_1, \bar{v}_2 \in \Delta([M])$, 
    \begin{align}
        \TV \left(\PP(\iota | \bar{v}_1), \PP(\iota | \bar{v}_2) \right) \ge \frac{\alpha}{2} \|\bar{v}_1 - \bar{v}_2\|_1.  
    \end{align} 
\end{assumption}
With these definitions at hand, we can define the setting studied in this work. We consider the LMDP with Prospective Side Information, which we refer to as LMDP-$\Psi$, which is the tuple $\theta = (\mathbb{I}, \{p_m, \mathbb{T}_m, \mathbb{O}_m\}_{m=1}^M) \in \Theta$. %\yecomment{there's a redundancy here since we need to discuss about the optimal policy again} 

Accordingly, our goal is now to learn an $\epsilon$-optimal policy from a larger class of policies $\Pi: \mI \times (\mA \times \mO \times \mS)^* \rightarrow \Delta(\mA)$ that exploits the prospective side information, given prior to each episodic interaction. An important subclass of this larger policy class is the {\it side information blind policies} $\Pi_{\texttt{blind}}: (\mA \times \mO \times \mS)^* \rightarrow \Delta(\mA)$, that does not exploit the prospective side information within each trajectory. As we see, the nature of the problem becomes different by the capacity of the policy class.

%We consider a policy class $\Pi: (\mA \times \mO \times \mS)^* \rightarrow \Delta(\mA)$ which contains all policies history dependent policies, i.e., that map all past observations from each time step to an action. \yecomment{formally, the optimal policy also depends on the side information. we may want to move the definition to later so we can refer to the side information.}
% \yecomment{next is the definition of the optimal policy, not a near optimal policy}

The optimal policy $\pi^*$ is defined as the optimal history-dependent policy that maximizes the expected cumulative reward 
$$V^\star = \max_{\pi \in \Pi} V^\pi := \Exs^\pi \left[ \tssum_{t=1}^H r_t( o_t) \right],$$
where the expectation is taken over latent contexts and rewards generated by an LMDP instance, following policy $\pi$. We let $\pi^*_{\texttt{blind}}$ be the counterpart in the smaller policy class~$\Pi_{\texttt{blind}}$.

% \yecomment{this paragraph shouldnt be here} As mentioned earlier, without further assumptions on the model, the problem is known to require at least $\Omega(SA)^M/\epsilon^2$ episodes to learn a near-optimal policy in the worst case. Therefore, with the growing number of contexts $M=\omega(1)$, it is essential to consider additional information that we can exploit to break the exponential barrier. 

\paragraph{Notation} We occasionally use the symbol $\lesssim$ to mean that the inequality holds up to some absolute constant. We use $\lesssim_{\texttt{P}}$ when it holds up to some {\it problem dependent} polynomial factors. To simplify notation, we occasionally denote pair-wise quantities as $x_t := (s_t,a_t)$, $y_t := (o_t, s_{t+1})$. $\Ber(p)$ denotes a Bernoulli random variable with parameter $p \in [0,1]$. For arbitrary full column-rank matrix $M$, $M^{\dagger}$ is a left-inverse of $M$ such that $M^{\dagger} M = I$.

\section{Related Work}

The study of learning algorithms for LMDPs was initiated within the framework of long-horizon multitask RL  \cite{taylor2009transfer, brunskill2013sample, hallak2015contextual, liu2016pac}, where full information on the latent contexts is revealed for a long-enough episode. However, problems in which full information on the latent context is not revealed cannot be solved through this framework. \citet{kwon2021rl} considered the sample complexity of learning a near-optimal policy for LMDPs without any assumptions. Unfortunately, their lower bound is exponential in the number of contexts, even when the transition dynamics are shared \cite{kwon2023reward, kwon2022tractable}. Hence, further investigation on the natural assumption for which LMDPs are efficiently learnable is required. To overcome the fundamental barriers in LMDPs, a few works have considered the assumption of giving true information in hindsight \cite{kwon2021rl, zhou2022horizon, lee2023learning}, as discussed earlier.

%\yecomment{we should stick with LMDPs instead of Latent MDPs(changed in places i've seen)} %To overcome the fundamental barriers in LMDPs, a few works have considered the assumption of giving true information in hindsight \cite{kwon2021rl, zhou2022horizon, lee2023learning}. Side-information can be considered as  Note that if the true information is given at the beginning, the problem becomes fully observable. In contrast, the side information does not fully resolve the partial observability of latent states even if it is given at the beginning. \yecomment{we discussed about that in the intro: should this be removed? or maybe just mentioned in a single sentence}

%-  \jycomment{maybe here why it is not direct from their results...} \yecomment{i think explaning this is really imporatant and deserves its own subsection }

Another related line of work is concurrent multitask learning \cite{hu2021near, maillard2014latent, gentile2017context, kwon2022coordinated}. Considering the label of each task as side information, this setting can be viewed as special cases of LMDPs with rich side information, analogous to rich observation in Block MDPs \cite{krishnamurthy2016pac, zhang2022efficient}. However, in these works the latent MDP is decodable from the observations at the beginning of each trajectory. Hence, this setting does not capture challenges that arise due to partial observability, when the latent state is not decodable. 

Another related work to our setting is the {\it multi-step} weakly revealing POMDP, where an agent must play sub-optimal actions to obtain weakly-revealing information \cite{golowich2022learning, liu2023optimistic, chen2023lower}. In this setting, a similar lower bound of $K^{2/3}$ regret has been reported in~\citet{chen2023lower}. While our lower bound construction is partially inspired by theirs, the LMDP-$\Psi$ setting is different since we obtain the weakly-revealing information ``for free'' at the beginning of each episode.  %\jycomment{edit}

% In this case, the $\sqrt{K}$-regret with a better dependence on problem parameters can be achieved \cite{hu2021near}. In contrast, we prove $K^{2/3}$ regret lower-bound for Latent MDPs with prospective side information, showing a fundamental difference from these lines of work.
\paragraph{LMDP-$\Psi$ is not a Weakly Revealing POMDP.}

% \yecomment{Made some edits in the next paragraph}
The recent line of work on weakly revealing POMDPs \cite{liu2022partially, liu2023optimistic, uehara2022provably, chen2022partially, chen2023lower} is the most closely related to ours. 
% From broader perspective, there has been substantial progress in learning weakly-revealing POMDPs, closely related setting to ours . %The main assumption required in these works is the existence of a {\it known} set of sufficient future tests, which (weakly) reveals the information of underlying belief states {\it in hindsight}. 
% In particular, under the weakly revealing assumption and further technical assumptions the results of \citet{liu2022partially, chen2022partially} shows that $\sqrt{K}$-regret is possible. 
Next, we elaborate on the differences between the settings. These highlight both the novelty and challenges in tackling the LMDP-$\Psi$ problem.

% In Section~\ref{subsection:why_not_POMDP} we show that, surprisingly, LMDP with prospective side information is not included within the class of weakly revealing POMDPs. Further, Table~\ref{table:regret_result} reveals technical differences between the settings. The lower bound for LMDPs with prospective side information establishes that the regret is lower bounded by $\Omega(K^{2/3})$ or suffers from exponential dependence in $M$. This form is substantially different than the upper bound for weakly revealing POMDPs.
% Our setting deviates from this line of work in that (i) when conditioned on the prospective side information, the remaining future may not be useful in hindsight, and (ii) even though the weakly-revealing information is given for free, $\sqrt{K}$-regret (where $K$ is the number of episodes) is not possible within a polynomial number of interactions. 

% \paragraph{LMDP-$\Psi$ are not Weakly Revealing POMDPs.}
% \label{subsection:why_not_POMDP}
% % \yecomment{would be good to be more precise here. Add discussion about the difference between POMDPs and prospective side information: $P(i_t \mid s_t,historical_observations) \neq P(i_t \mid s_t)$ unlike standard POMDP modeling assumption (thinking about $i_t$ as an observation we see at each time step).}

% We elaborate on key differences between the LMDP-$\Psi$ and the weakly revealing POMDP setting. These highlights both the novelty and challenges tackling the LMDP-$\Psi$ setting.

\begin{itemize}
    \item \textit{ Standard POMDP modeling assumptions are violated in the presence of prospective information.} For the LMDP-$\Psi$ setting, the available observations between different time step are not independent, conditioned on the latent state. Let the available observation at each time step be $\tilde{o}_t:=(o_t,\iota)$,  {\it i.e.}, a combination of the observation and the available initial prospective side information. Trivially, the common conditional independence on the latent state assumption for the observation generation process does not hold. It does not necessarily hold that $\PP^\pi(\tilde{o}_t \mid s_t,m) \neq \PP^\pi( \tilde{o}_t \mid s_t, m, \tilde{o}_{t-1})$: $\tilde{o}_{t-1}$ contains information on $\tilde{o}_{t}$ since the prospective information, $\iota$, is fixed during an episode. That is, there is a non-trivial correlation between observations. Unlike LMDP-$\Psi$, in the common setting POMDP setting, and weakly revealing POMDPs~\cite{liu2022partially}, the observation is independent of historical information conditioning on the latent state.
    \item \textit{Regret guarantees are fundamentally different.} As depicted in Table~\ref{table:regret_result}, the regret lower bound for LMDP-$\Psi$, without the exponential on the number of latent contexts, is $\Omega(K^{2/3})$. Such a lower bound is fundamentally different than the $O(\sqrt{K})$ upper bound for weakly revealing POMDPs. This highlights a key difference between the settings established by our results. 
\end{itemize}

\section{Learning in LMDP-$\Psi$}
In this section, we present our algorithmic results as well as lower bound analysis.

\subsection{Warm Up: $\sqrt{K}$-Regret within $\Pi_{\texttt{blind}}$}

Consider the problem of learning a near-optimal policy only in the blind policy class $\Pi_{\texttt{blind}}$. Such a setting is equivalent to the one in which the prospective side information is provided in hindsight, and thus, the problem falls into the setting of well-conditioned PSR studied in~\citet{liu2023optimistic}. To see this, define problem operators $B(o,s_{+1}|s,a) = \mathbb{I} \cdot \diag ([\PP(o,s_{+1} | m,s,a)]_{m=1}^M) \cdot \mathbb{I}^{\dagger}$ and $b_0 = \mathbb{I} w$. We can easily verify that for any blind policy $\pi \in \Pi_{\texttt{blind}}$ and trajectory $\tau = (s_1, a_1, o_1, ..., s_H, a_H, o_H)$, 
\begin{align*}
    \PP^\pi (\iota, \tau) = \bm{e}_\iota^\top \cdot \Pi_{t=1}^{H} B {(o_t, s_{t+1} |s_t,a_t)} \cdot b_0 \cdot \pi(\tau),
\end{align*}
where $\pi(\tau) = \Pi_{h=1}^H \pi(a_h | s_{1},...,s_{h})$. We define $s_{H+1} := \emptyset$ in the above expression. Let 
\begin{align}
    &\omega_t := (r_{t}, s_{t+1}, a_{t+1}, ..., r_H), \nonumber \\
    &\psi (\omega_t, \iota | s_t, a_t)^\top := \bm{e}_{\iota}^\top \cdot \Pi_{h=t}^{H} B {(o_{h},s_{h+1}|s_{h},a_{h})}, \label{eq:define_futures_notation}
\end{align}
where $\omega_t$ is the future partial trajectory from time step $t$. With this, the system reparameterized by $B$ and $b_0$ with the blind policy class is a well-conditioned PSR, as defined in~\citet{liu2023optimistic} (see their Condition 4.3), {\it i.e.,} for any $t \in [H]$ and any policy $\pi\in \Pi_{\texttt{blind}}$ it holds that
\begin{align}
    \max_{b: \|b\|_1=1} \sum_{\iota, \omega_t} \pi(\omega_t) |\psi(\omega_t, \iota | s_t, a_t)^\top b| \le \frac{M}{\alpha}. \label{eq:well_conditioned_PSR}
\end{align}
%This can be shown to hold for any policy independent of the history before step $t$ \yecomment{add a reference to the appendix?}.
With the above condition, since no extra tests are required to obtain $\iota$, this allows us to apply the Optimistic-MLE (O-MLE) algorithm introduced in \citet{liu2022partially} for regret minimization (see Algorithm \ref{algo:psr_ucb_blind}). 
\begin{algorithm}[t]
    \caption{Regret Minimization within $\Pi_{\texttt{blind}}$}
    \label{algo:psr_ucb_blind}
    \begin{algorithmic}[1]
        \STATE{Initialize $\mD^0 = \emptyset$, $\mC^0 = \Theta$}
        \FOR{$k = 1 ... K$}
            \STATE{\color{blue}{\# Optimistic Policy Search}}
            \STATE{Pick $(\theta^k, \pi^k) = \arg\max_{\theta \in \mC^k, \pi \in \Pi_{\texttt{blind}}} V^{\pi^k}_{\theta^k}$}
            \STATE{Get $\tau^k = (s_1^k, a_1^k, ..., r_H^k), \iota^k$ by executing $\pi^k$}
            \STATE{\color{blue}{\# Confidence Set Construction}}
            \STATE{$\mD^k \leftarrow \mD^{k-1} \cup \{(\iota^k, \tau^k, \pi^k)\}$ and update $\mC^k$ using \eqref{eq:mle_confidence}}
        \ENDFOR
    \end{algorithmic}
\end{algorithm}

We can follow the analysis of the optimistic-MLE approach for well-conditioned PSRs \cite{liu2023optimistic}, yielding the following theorem:
\begin{theorem}
    \label{theorem:regret_upper_bound_blind}
    Let $\pi_{\texttt{blind}}^*$ be the optimal policy in $\Pi_{\texttt{blind}}$ for the true environment $\theta^*$. With probability greater than $1 - \delta$, the regret of Algorithm \ref{algo:psr_ucb_blind} (with respect to the optimal blind policy) satisfies
    \begin{align*}
        \sum_{k=1}^K V_{\theta^*}^{\pi_{\texttt{blind}}^*} - V_{\theta^*}^{\pi^k} \lesssim \frac{M^{3/2} H^2}{\alpha} \sqrt{SAK \log(|\Theta|/\delta) (\log K)}. 
    \end{align*}
\end{theorem}
Note that the size of model class $|\Theta|$ is typically exponential in the number of free parameters that define the system, and we would hope to bound the regret with a $\log |\Theta|$ term for general function classes. For the tabular case with finite supported observation and prospective side information, this term scales as $\log |\Theta| = \tilde{O}(M(S^2A + SA|\mO|) + M|\mI|)$. 
% The more interesting regime is when the absolute cardinality $|\mO|$ or $|\mI|$ is arbitrarily large ({\it e.g.,} quantized over continuous space), for this case, but the regret only scales with the limited capacity of model class $\log |\Theta|$.

\subsection{What's Wrong with $\pi_{\texttt{blind}}^*$?}

Even if we obtain a sublinear $O(\sqrt{K})$-regret compared to $\pi^*_{\texttt{blind}}$, note that the original goal is to learn the true optimal policy $\pi^* \in \Pi$ which exploits the prospective side information within each trajectory. Therefore, the notion of true regret must be defined in a stronger sense: 
\begin{align}
    \text{Regret}(K) = \tssum_{k=1}^K V_{\theta^*}^{\pi^*} - V_{\theta^*}^{\pi^k}. \label{eq:stronger_regret_bound}
\end{align}
The overall measure of performance should be on obtaining $\sqrt{K}$-regret with the above stricter definition. 

Another issue is, by converting the argument of regret-minimization to sample-complexity, we can obtain $\epsilon$-optimal policy from Algorithm \ref{algo:psr_ucb_blind} with $\epsilon = O(1/\sqrt{K})$. However, a naive conversion of near-optimal policies in $\Pi_{\texttt{blind}}$ would only guarantee $(|\mI| \epsilon)$-optimality for the larger class of policies $\Pi$. To see this, suppose O-MLE returns a model $\theta$ such that for all $\pi \in \Pi_{\texttt{blind}}$,
\begin{align*}
    \TV \left(\PP_{\theta}^{\pi}(\iota, \tau) , \PP_{\theta^*}^{\pi}(\iota, \tau) \right) \le \epsilon,
\end{align*}
%for the model $\theta^k$, estimated by the O-MLE algorithm, and the corresponding policy it interacts with the environment in the $k^{th}$ episode. 
For the individual $\iota$, however, we can only infer in the worst case that 
\begin{align*}
    \PP_{\theta^*}(\iota) \cdot \TV \left(\PP_{\theta}^{\pi}(\tau|\iota) , \PP_{\theta^*}^{\pi}(\tau|\iota) \right) \le \min(\PP_{\theta^*}(\iota), \epsilon).
\end{align*}
Thus, when considering a larger policy class $\pi \in \Pi$, a naive analysis would lead to the following upper bound%, \yecomment{not quite precise, try and rephrase?}
\begin{align*}
    &\sum_{\iota} \PP_{\theta^*}(\iota)  \TV \left(\PP_{\theta^*}^{\pi(\cdot | \iota)}(\tau | \iota), \PP_{\theta^k}^{\pi(\cdot|\iota)}(\tau | \iota) \right) \le \min(1, |\mI| \epsilon),
\end{align*}
since for every $\iota$ we use different policy $\pi(\cdot | \iota)$, but a naive analysis would result in a loose bound with multiplicative amplification of the error. Since we consider a large or (almost) continuous observation, the result should not directly depend on $|\mI|$, and, instead depend on $\log(|\Theta|)$.

% \paragraph{Can O-MLE Simply Learn with $\Pi$?} Instead of choosing a policy from $\Pi_{\texttt{blind}}$, we may consider picking a policy to execute from $\Pi$ in Algorithm \ref{algo:psr_ucb_reward_free}. However, once we exploit $\iota$ at the beginning, the LMDP-$\Psi$ system is not a well-conditioned PSR according to the definition in \eqref{eq:well_conditioned_PSR}. That is, once $\iota$ is a part of history (and not a part of the future), we cannot find a way of defining future operators \eqref{eq:define_futures_notation} that satisfies the condition \eqref{eq:well_conditioned_PSR}, as discussed in \ref{subsection:why_not_POMDP}. \yecomment{this paragraph is a bit confusing (also, should this be stated in section 2.1?)}

\subsection{Hardness of $\sqrt{K}$-Regret}
The first question with prospective side information is whether we can still achieve $\sqrt{K}$-regret in the stronger sense of equation~\eqref{eq:stronger_regret_bound}, i.e., to achieve guarantees with respect to a stronger notion of optimal policy. Surprisingly (and rather disappointingly), when learning with a larger policy class with the stronger notion of regret, we show that it is impossible to obtain $\sqrt{K}$-regret unless $K$ is larger than $A^{\Omega(M)}$.

\begin{theorem}
    \label{theorem:regret_lower_bound}
    There exists a family of LMDP-$\Psi$s, $\Theta_{\text{hard}}$, and a reference model $\theta_0$ with $\alpha$-prospective side information, such that for any algorithm, the regret of the worst-case instance satisfies with $\alpha < 1 / (256\sqrt{M})$, 
    % \yecomment{should set $\alpha$ smaller than some constant?}
    {\ifarxiv
        \begin{align*}
            \inf_{\psi: \texttt{Algs}} &\sup_{\theta \in \Theta_{\text{hard}} \cup \{\theta_0\}} \sum_{k=1}^K V_{\theta}^{\pi^*} - V_{\theta}^{\pi^k(\psi)} \gtrsim \min \left( \frac{(A/3)^{(M/4)}}{M\epsilon}, \frac{A}{M \alpha^2\epsilon^2}, \frac{K\epsilon}{M} \right).
        \end{align*}
    \else
    \begin{align*}
        \inf_{\psi: \texttt{Algs}} &\sup_{\theta \in \Theta_{\text{hard}} \cup \{\theta_0\}} \sum_{k=1}^K V_{\theta}^{\pi^*} - V_{\theta}^{\pi^k(\psi)} \\
        &\gtrsim \min \left( \frac{(A/3)^{(M/4)}}{M\epsilon}, \frac{A}{M \alpha^2\epsilon^2}, \frac{K\epsilon}{M} \right).
    \end{align*}
    \fi}
\end{theorem}
By optimizing over $\epsilon$, we obtain the following lower bound:
\begin{corollary}
    The regret of any algorithm for the worst-case family of instances satisfy
    {\ifarxiv
        \begin{align*}
            \inf_{\psi: \texttt{Algs}} &\sup_{\theta \in \Theta_{\text{hard}} \cup \{\theta_0\}} \sum_{k=1}^K V_{\theta}^{\pi^*} - V_{\theta}^{\pi^k(\psi)} \gtrsim_\texttt{P} \min \left( A^{\Omega(M)} \sqrt{K}, \left(\frac{A}{\alpha^2} \right)^{1/3} K^{2/3}) \right).
        \end{align*}
    \else
    \begin{align*}
        \inf_{\psi: \texttt{Algs}} &\sup_{\theta \in \Theta_{\text{hard}} \cup \{\theta_0\}} \sum_{k=1}^K V_{\theta}^{\pi^*} - V_{\theta}^{\pi^k(\psi)} \\
        &\gtrsim_\texttt{P} \min \left( A^{\Omega(M)} \sqrt{K}, \left(\frac{A}{\alpha^2} \right)^{1/3} K^{2/3}) \right).
    \end{align*}
    \fi}
\end{corollary}
This lower bound implies the impossibility of designing a learning algorithm with $\poly(M)\sqrt{K}$-regret. Instead, next, we aim to derive an algorithm with an  upper bound of  $\poly(M)K^{2/3}$ on its regret, i.e., a regret guarantee with no exponential dependence in the number of latent contexts.

\subsection{Pure Exploration within $\Pi_{\texttt{blind}}$ is Sufficient}
%\yecomment{maybe state here  (thing it is more precise and easier to understand).} 
In this section, we present an explore-then-exploit strategy that the optimal $O(K^{2/3})$ regret. When Algorithm~\ref{algo:psr_ucb_blind} (or a reward-free version of it) terminates, the guaranteed inequality is usually on the total variation distance between any model in the confidence set $\theta \in \mC^K$ and true models $\theta^*$:
\begin{align*}
    \max_{\pi \in \Pi_{\texttt{blind}}} \TV(\PP^{\pi}_{\hat{\theta}}, \Pi^{\pi}_{\theta^*}) \le \epsilon. 
\end{align*}
As discussed earlier, this is not sufficient, and we need a stronger notion of termination criterion, which ensures that all reachable {\it belief (and the PSR) states} have been sufficiently explored in all models in the remaining confidence. Formally, define the reward bonus for any history at a state-action pair $x := (s,a)$ as
\begin{align*}
    \hat{\Lambda}_{t}^k (x) &= \lambda_0 I + \sum_{j < k} \indic{x_t^{j} = x} \bar{b}_{\theta^k} (\tau_t^{j}) \bar{b}_{\theta^k} (\tau_t^{j})^\top, \\
    \tilde{r}^k(\tau_t) &= \| \bar{b}_{\theta^k} (\tau_t) \|_{\hat{\Lambda}_{t}^k(x)^{-1}},
\end{align*}
where $\bar{b}_{\theta}(\tau_t) = \frac{b_{\theta}(\tau_t)}{\|\bar{b}_{\theta}(\tau_t)\|_1}$ is a normalized PSR of history $\tau_t$ in a model $\theta$ when a {\it blind} policy is executed. The key observation is, when considering a larger class of policies $\Pi$, we can show that
\begin{align}
    \max_{\pi \in \Pi} \TV(\PP^{\pi}_{\hat{\theta}}, \PP^{\pi}_{\theta^*}) &\lesssim_{\texttt{P}} \max_{\pi \in \Pi_{\texttt{blind}}} \Exs_{\hat{\theta}}^{\pi} \left[\tssum_{t=1}^H \tilde{r}^k (\tau_t) \right]. \label{eq:connect_pi_and_pi_blind}
\end{align}

\begin{algorithm}[t]
    \caption{Pure Exploration for LMDP-$\Psi$}
    \label{algo:psr_ucb_reward_free}
    {{\bf Input:} Termination condition $\epsilon_{\texttt{pe}} := \frac{\alpha \epsilon}{10 H M^2 \sqrt{\lambda_0 M^2/\alpha^2 + \beta}}$, Regularizer $\lambda_0 := \frac{\beta M^2H^2}{\alpha^2}$}
    \begin{algorithmic}[1]
        \STATE{Initialize $\mD^0 = \emptyset$, $\mC^0 = \Theta$}
        \FOR{$k = 0 ... K-1$}
            \STATE{\color{blue}{\# Execute the Worst Blind Policy}}
            \STATE{Pick any $\theta^k \in \mC^k$}
            \STATE{$\pi^k = \arg\max_{\pi \in \Pi_{\texttt{blind}}} \tilde{V}_{\theta^k, \tilde{r}^k}^\pi$}
            \STATE{\textbf{If} $\tilde{V}_{\theta^k, \tilde{r}^k}^\pi \le \epsilon_{\texttt{pe}}$, then \textbf{break} }
            \STATE{Get $\iota^k$ and $\tau^k = (s_1^k, a_1^k, ..., r_H^k)$ by executing $\pi^k$}
            \STATE{\color{blue}{\# Confidence Set Construction}}
            \STATE{$\mD^k \leftarrow \mD^{k-1} \cup \{(\tau^k, \iota^k, \pi^k)\}$ and update $\mC^k$ using \eqref{eq:mle_confidence}}
        \ENDFOR
        \STATE \textbf{return} $\hat{\theta} = \theta^k$
    \end{algorithmic}
\end{algorithm}

A recent result of \citet{huang2023provably} (see their Lemma 6) gives an explicit bound on the quantity $\Exs_{\hat{\theta}}^{\pi} \left[\tssum_{t=1}^H \tilde{r}^k (\tau_t) \right]$, instead of bounding the total-variation distance indirectly from the elliptical potential lemma. Therefore, their pure exploration algorithm, but {\it only within a class of blind policies} $\Pi_{\texttt{blind}}$, is sufficient to learn the optimal policy in a larger class of policy $\Pi$. We mention that before the result of \citet{huang2023provably}, direct bound on the cumulative bonus of trajectories did not exist.

Formally, we consider Algorithm \ref{algo:psr_ucb_reward_free}, where we let $\tau_{t} := (s_1, a_1, ..., s_t, a_t)$ be a partial trajectory up to time-step $t$ without prospective side information. The expected cumulative bonus at the $k^{th}$ episode in the empirical model is defined as
\begin{align*}
    \tilde{V}_{\theta^k, \tilde{r}^k}^\pi := \Exs_{\tau \sim \PP_{\theta^k}^\pi} \left[\tssum_{t=1}^H \tilde{r}^k (\tau_t) \right].
\end{align*}
The confidence set is given based on the likelihood of each model:
\ifarxiv
    \begin{align}
        \label{eq:mle_confidence}
        \mC^k := \Big\{ &\theta \in \Theta \ \Big| \tssum_{(\iota, \tau, \pi) \in \mD^k} \log \PP^{\pi}_{\theta} (\iota, \tau) \ge \max_{\theta' \in \Theta} \tssum_{(\iota, \tau, \pi) \in \mD^k} \log \PP^{\pi}_{\theta'} (\iota, \tau) - \beta \Big\}. 
    \end{align}
\else
\begin{align}
    \label{eq:mle_confidence}
    \mC^k := \Big\{ &\theta \in \Theta \ \Big| \tssum_{(\iota, \tau, \pi) \in \mD^k} \log \PP^{\pi}_{\theta} (\iota, \tau)\nonumber \\
    &\ge \max_{\theta' \in \Theta} \tssum_{(\iota, \tau, \pi) \in \mD^k} \log \PP^{\pi}_{\theta'} (\iota, \tau) - \beta \Big\}. 
\end{align}
\fi
$\beta$ is pre-defined by the concentration of likelihood value, and is given by $\log (K |\Theta| / \delta)$ as shown in Lemma \ref{lemma:mle_traj_concentration}. Note that from the construction of the confidence set $\mathcal{C}^k$, for all $k \in [K]$, we know that with probability at least $1- \delta$, 
\begin{align*}
    - \sum_{(\tau,\pi) \in \mD^k} \log \left( \frac{\PP^\pi_{\theta^k} (\iota, \tau)}{\PP^\pi_{\theta^*} (\iota, \tau)} \right) \le 2 \beta.
\end{align*}
Thus, we may simply choose the maximum likelihood estimator (MLE). We obtain the following guarantee:
\begin{theorem}
    \label{theorem:reward_free_exploration}
    Let $\epsilon_{\texttt{pe}}, \lambda_0$ as defined in the input in Algorithm \ref{algo:psr_ucb_reward_free}. Then, with probability at least $1 - \delta$, Algorithm \ref{algo:psr_ucb_reward_free} returns a model $\hat{\theta}$ after at most $K$ episodes where
    \begin{align}
        K = O\left( \frac{M^{8} H^4 SA \cdot \log (K|\Theta|/\delta) \log (K)}{\alpha^6 \epsilon^2} \right), \label{eq:K_sample_complexity}
    \end{align}
    Furthermore, the optimal policy $\pi^*_{\hat{\theta}} \in \Pi$ for the returned model $\hat{\theta}$ is an $\epsilon$-optimal policy for $\theta^*$ with probability at least $1 - \delta$, {\it i.e.,} $\left| V_{\theta^*}^{\pi^*} - V_{\theta^*}^{\pi^*_{\hat{\theta}}} \right| \le \epsilon$.
\end{theorem}
Finally, the sample complexity guarantee can naturally be converted into a regret guarantee by a standard explore-then-exploit approach.  That is, by playing $\epsilon^{-2} = O(K^{2/3})$ to obtain an $\epsilon$-optimal policy and exploit the learned policy for the remaining episode. For regret minimization, we get:
\begin{align*}
    \sum_{k=1}^K V_{\theta^*}^{\pi^*} - V_{\theta^*}^{\pi^k} \lesssim \left(\frac{M^{8} H^4 SA \cdot \log(|\Theta| / \delta)}{\alpha^6} \right)^{1/3} K^{2/3},  
\end{align*}
regret bound up to logarithmic factors for $K$ episodes.

\section{Analysis}
In this section, we provide the upper and lower bounds proofs and intuition. 

\subsection{Upper Bound}
\label{subsec:upper_bound_overview}
Here, we provide the overview of analyzing Algorithm \ref{algo:psr_ucb_reward_free}. The main step is to establish the inequality of equation~\eqref{eq:connect_pi_and_pi_blind}. We adopt the idea from \citet{huang2023provably} of separating the concentration argument (for bounding the sum of TV distances) and the elliptical potential argument. In addition to the notation defined in equation~\eqref{eq:define_futures_notation}, we let 
\begin{align*}
    &b(\tau_{t}) := \Pi_{h=1}^{t-1} B {(o_{h}, s_{h+1}|s_{h},a_{h})} b_0, \\
    &\pi(\iota, \tau_t) := \Pi_{h=1}^{t} \pi(a_{h} | \iota, s_1, ..., s_{h}), \\
    &\pi(\omega_t | \iota, \tau_t) := \Pi_{h=t+1}^{H} \pi(a_{h} | \iota, s_1, ..., s_{h}).
\end{align*}

Our crucial observation on exploiting the prospective weakly revealing side information is the following conditional, on the value of $\iota$, well conditioning of the LMDP-$\Psi$ system:
\begin{lemma}
    \label{lemma:conditional_well-conditioned}
    Fix any prospective side information $\iota \in \mI$. For all $x_t = (s_t,a_t) \in \mS\times\mA$, $t \in [H]$, and $\pi$ that is independent of the history before time-step $t$, we have
    \begin{align*}
        \max_{b:\|b\|_1=1}\max_{\pi} \sum_{\omega_t} \pi(\omega_t) |\psi(\omega_t,\iota | x_t)^\top b| \le \frac{M}{\alpha}  \max_{m \in [M]} \PP(\iota|m). 
    \end{align*}
\end{lemma}
On the other hand, following the standard algebra to bound the total variation distance, we can bound $\TV(\PP_{\theta^*}^\pi, \PP_{\theta}^\pi)$ as for all $\theta$ as follows:
\begin{align*}
    \TV(\PP_{\theta^*}^\pi, \PP_{\theta}^\pi) &\le \sum_{t=1}^H \sum_{\iota,\tau_t} \pi(\tau_t|\iota) \sum_{\omega_{t}} \left| f(\omega_t, \iota) b_{\theta}(\tau_{t}) \right|,
\end{align*}
where $f(\omega_t, \iota) := \pi(\omega_t | \tau_t, \iota) \cdot \psi_{\theta^*}(\omega_{t+1}, \iota| x_{t+1})^\top$ $\left(B_{\theta^*} (y_t|x_t) - B_{\theta} (y_t|x_t) \right)$ is the term involving the operator difference between two models. The inner summation over the partial future trajectory $\omega_t$ can be split into the multiplication of the concentration error in PSR {\it conditioned} on~$\iota$: 
\begin{align*}
    \|\tssum_{\omega_t} f(\omega_t, \iota) \|_{\hat{\Lambda}_{t} (x_t)}
\end{align*}
and the cumulative sum of trajectory bonuses when the prospective side information $\iota$ is ignored: 
\begin{align*}
    \|\bar{b}_{\theta}(\tau_t)\|_{\hat{\Lambda}_{t} (x_t)^{-1}}. 
\end{align*}
For the concentration error in PSRs, we can apply the conditional concentration of total-variation distances for likelihood estimators (see Appendix \ref{lemma:concentration_conditioanl_tv_dist}) and Lemma \ref{lemma:conditional_well-conditioned}. For the cumulative bonuses, we use the termination condition of Algorithm \ref{algo:psr_ucb_reward_free}. Combining the two separate arguments, we can prove Theorem \ref{theorem:reward_free_exploration}. See Appendix \ref{appendix:proof:upper_bound} for the complete proofs.

\subsection{Lower Bound}
\ifarxiv
\begin{figure}
    \centering    
    \begin{subfigure}{.8\linewidth}
        \centering
        \includegraphics[width=.75\textwidth]{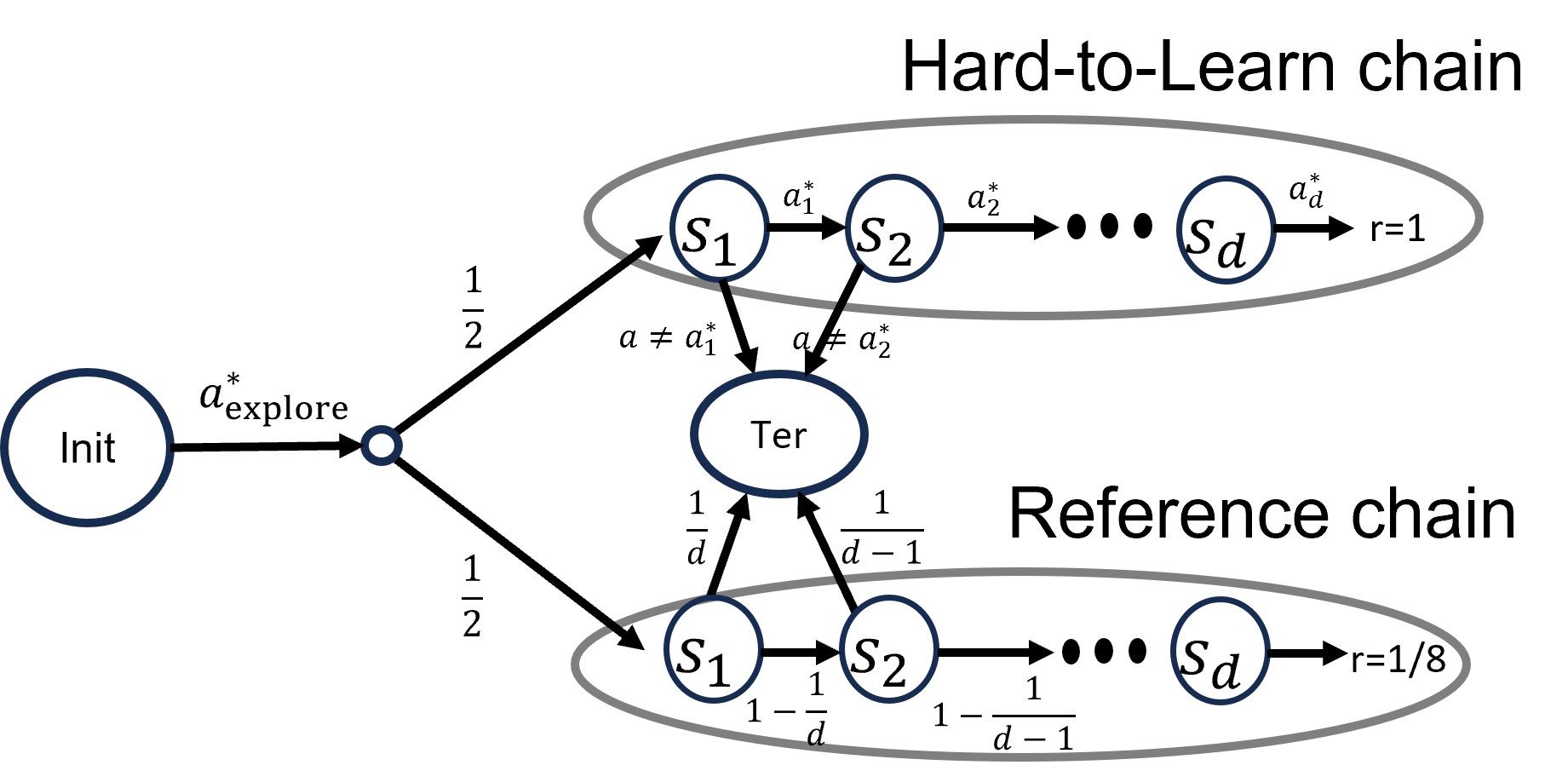}
        \caption{Case I: $\iota = \iota_{\text{hard}}$ does not reveal anything. The probability to get $\iota = \iota_{hard}$ is larger than $1/4$.}
    \end{subfigure}
    
    \begin{subfigure}{.8\linewidth}
        \centering
        \includegraphics[width=.5\textwidth]{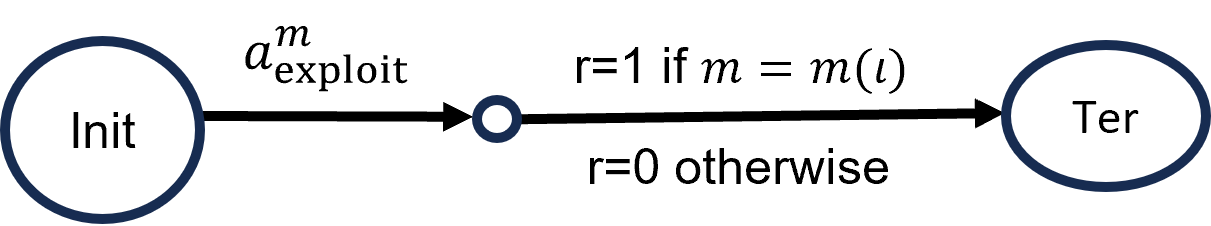}
        \caption{Case II: $\iota \neq \iota_{\text{hard}}$ nearly specifies the true context. In this case, one context $m(\iota) \in [M/2+1, ..., M]$ has a strong prior, {\it i.e.,} $\PP(m(\iota) | \iota) \ge 1/2$.}
    \end{subfigure}
    \caption{Hard instance. Optimal behaviors is denoted with the set of actions $\{ a_i^{*}\}_{i=1}^d$. The numbers on the arrow represent the probability of transitions under the optimal policy. 
    % Actions on the arrow mean transition happens when the action satisfying the condition is taken 
    The symbol $a\neq \bar{a}$ next to an arrow means that all actions that are not $\bar{a}$ result with the described transition.} % \yecomment{better description of the figure. change the actions to correspond to the same notation as in the text. remove the circles. change the notation of the lower figure to the complementary event.}}
    \label{fig:hard_instance}
\end{figure}
\else
\begin{figure}
    \centering    
    \begin{tabular}{cc}
        \includegraphics[width=0.45\textwidth]{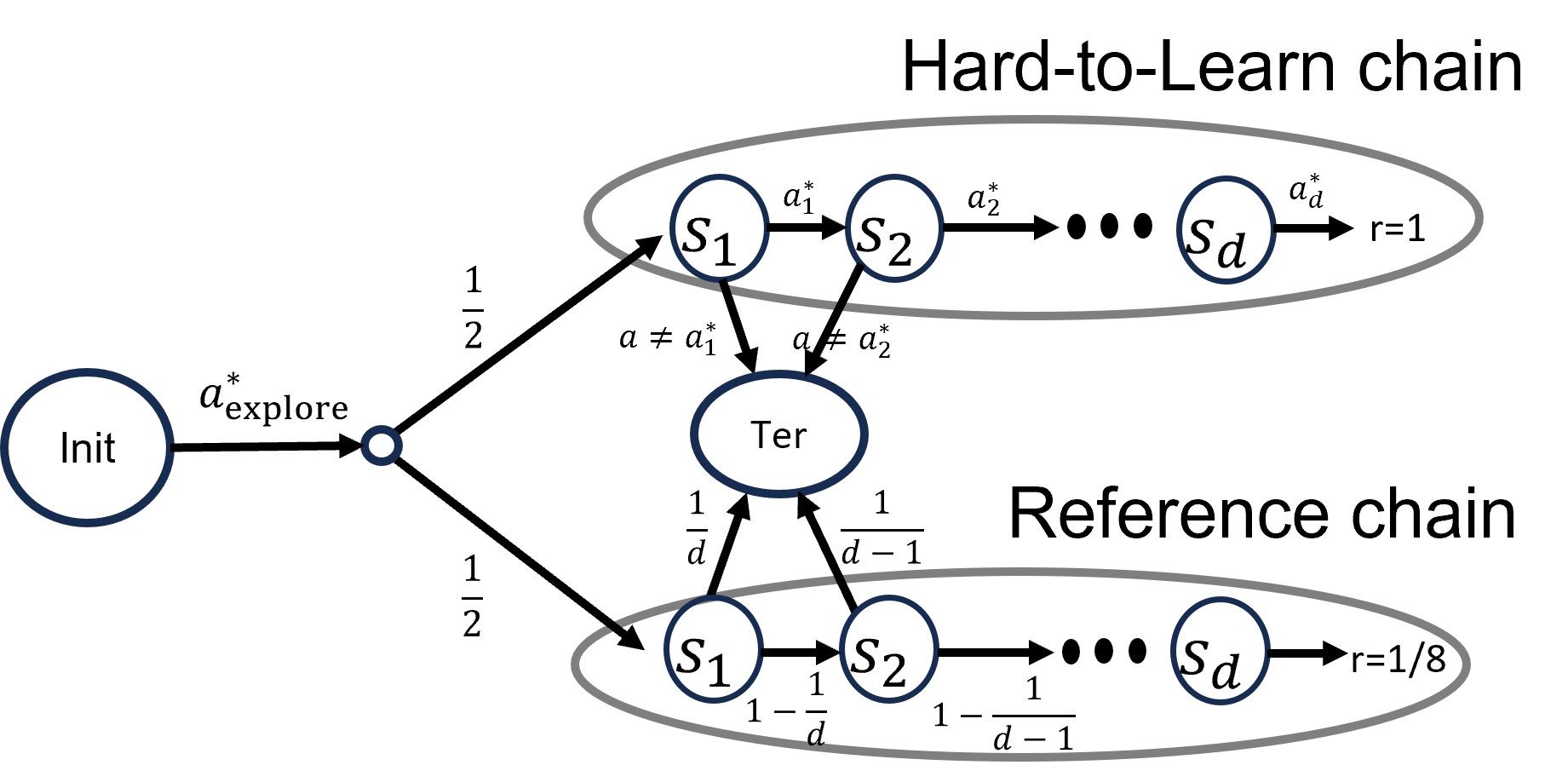} \\
        Case I: $\iota_{\text{hard}}$ does not reveal anything \\
        \\
        \includegraphics[width=0.45\textwidth]{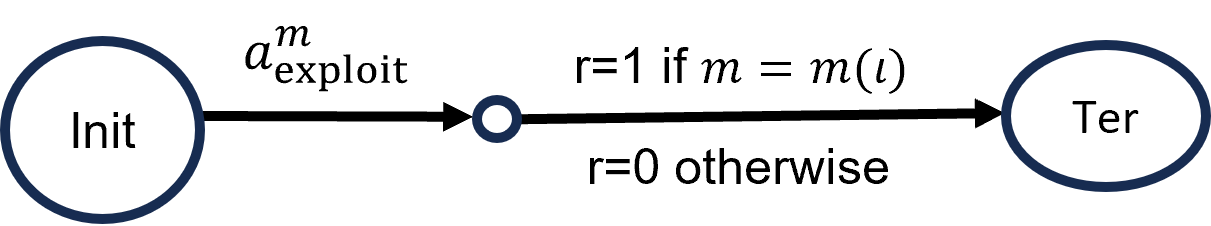} \\
        Case II: $\iota \neq \iota_{\text{hard}}$ nearly specifies the context
    \end{tabular}
    \caption{Optimal behaviors in the family of hard instances. The numbers on the arrow mean the probability of transitions under the optimal policy. Actions on the arrow mean transition happens when the action satisfying the condition is taken.}
    \label{fig:hard_instance}
\end{figure}
\fi

Next, we describe the lower bound construction and supply intuition for this result. Consider the following scenario (see Figure \ref{fig:hard_instance} for the class of LMDP-$\Psi$s): suppose that for a non-negligible portion of episodes, the prospective side information does supply any information on the latent context. That is, given the prospective side information $\iota_{\text{hard}}$ the posterior probability over the latent contexts is uniform, {\it i.e.,} $\PP(m | \iota_{\text{hard}}) = 1/M$, and $\iota_{\text{hard}}$ happens with constant probability, e.g., $1/4$. With $\iota_{\text{hard}}$ alone, however, learning the optimal action sequence $a_{1:d}^*$ (optimal policy) may suffer from an exponential lower bound $A^{\Omega(M)}$ since $\iota_{\text{hard}}$ supplies no information on the latent context. At the same time, playing any sub-optimal action sequence incurs an $\Omega(\epsilon)$ regret where $\epsilon$ is the required target accuracy.

On the other hand, any other prospective side information $\iota \neq \iota_{\text{hard}}$, provides a strong signal of one environment that is the most likely, {\it i.e.,} $\PP(m^*(\iota) | \iota) \ge 1/2$. Further, suppose that there is a unique exploiting action for each context that always gives a high reward, and playing any other action incurs $O(1)$-regret. For this environment, the regret of any algorithm is proportional to how many times the sub-optimal action is played when $\iota \neq \iota_{\text{hard}}$.

%To formalize the argument, suppose there are two chains of transition dynamics of length $d$, and one environment (or context) $\mM_1$ where we can expect higher-rewards when a correct action-sequence is taken. In the hard-to-learn chain, $s_{1:d}^{\text{hard}}$, only the correct length-$d$ action-sequence $a_{1:d}^{*}$ can pass through the chain and obtain a high reward, and otherwise it falls into the absorbing state with 0 reward. In the reference chain, $s_{1:d}^{\text{ref}}$, a transition at $s_t$ to the next state happens with probability $1 - \frac{1}{(d-t+1)}$ regardless of taken actions. However,  without side information, the two chains cannot be distinguished when the taken action sequence $a_{1:d}$ does not exactly match the correct sequence $a_{1:d}^*$. This results in the $\Omega(A^d)$ lower bound. 

However, it is still essential to learn the optimal sequence of actions $a_{1:d}^*$ in order to behave optimally under $\iota_{\text{hard}}$. Therefore, to avoid the exponential lower bound, we should be aided by good prospective side information $\iota \neq \iota_{\text{hard}}$ despite the strong signal of the underlying model. We can construct internal dynamics such that we need to explore the two chains for at least $\Omega(A/\alpha^2\epsilon^2)$ episodes to identify the optimal action sequence $a_{1:d}^*$ when $\iota \neq \iota_{\text{hard}}$. Combining these arguments, the regret lower bound should be at least $\min\left(\frac{A^\Omega(M)}{\epsilon}, \frac{A}{\alpha^2\epsilon^2}, K\epsilon\right)$, yielding Theorem \ref{theorem:regret_lower_bound}.

To obtain the multiplicative dependence on $\alpha$ and $\epsilon$, the actual construction of the hard instance family is slightly more complicated. We assume that $M$ is sufficiently large and a multiple of 4, and let $d = M/4$. We also assume that $\alpha \ll 1$ is a sufficiently small constant. We let the time step start from $t=0$ at the initial state $s_{\text{init}}$. Next, we describe the construction of the hard LMDP-$\Psi$ class:
\paragraph{State Space.} There are four categories of states. The initial state $s_{\text{init}}$, absorbing state $s_{\text{ter}}$ (which means essentially an episode is terminated), a chain of states constructing a hard-to-learn system $s^{\text{hard}}_{1:d}$, and another chain of states constructing a reference system $s^{\text{ref}}_{1:d}$. 

\paragraph{Action Space.} The set of actions at the initial time step consists of a set of candidate exploring actions $\mathcal{A}_{\text{explore}}$ and exploiting actions $\mA_{\text{exploit}} := \{a^m_{\text{exploit}}\}_{m=M/2+1}^{M}$ that control the dynamics at the initial state. The action set at time steps $1,\cdots, d$, denoted by $\mA_{\text{control}}$, controls the dynamics in hard-to-learn and reference chains of the system. At the initial time step, only one action of $\mathcal{A}_{\text{explore}}$ is a true exploring action $a_{\text{explore}}^*$. At time steps $1,\cdots,d$ only one action sequence $a_{1:d}^* \in \mA_{\text{control}}^{\bigotimes d}$ is the optimal sequence.

\paragraph{Latent Environments and Initial Dynamics.} There are three groups of MDPs: $\mG_{\texttt{learn}}$, $\mG_{\texttt{ref}}$, and $\mG_{\texttt{obs}}$. All MDPs always start from the same starting state $s_{\text{init}}$. 

$\mG_{\texttt{learn}}$ consists of $(M/4)$ MDPs, $\mM_1, ..., \mM_{M/4}$, which essentially form the hard to learn example from~\citet{kwon2021rl} when no prospective side information is provided. In any of these environments, in the beginning, when the `true' explore action $a_{\text{explore}}^*$ is played, it transitions to the starting of hard-instance chain $s^{\text{hard}}_{1}$ with some small probability.

Similarly, $\mG_{\texttt{ref}}$ consists of another $(M/4)$ MDPs, $\mM_{M/4+1}, ..., \mM_{M/2}$, and the purpose of $\mG_{\texttt{ref}}$ is to confuse the learning the optimal action sequence in the hard-to-learn chain, as we make the prospective side information hard to distinguish whether an MDP belongs to $\mG_{\texttt{learn}}$ or $\mG_{\texttt{ref}}$. More precisely, under $\iota_{\text{hard}}$, it is hard to identify which one is the hard-to-learn or reference chain, and thus it is hard to identify $a_{\text{explore}}^*$. This is crucial to build a multiplicative lower bound on $\alpha$ and $\epsilon$.

The rest of $(M/2)$ MDPs, indexed by $\mM_{M/2+1}, ..., \mM_M$, belong to the almost observable group $\mG_{\texttt{obs}}$. In each environment of this group $\mM_m \in \mG_{\texttt{obs}}$ where $m=M/2+1,...,M$, executing $a^{m}_{\text{exploit}}$ at the initial time step step results with a reward 1, and gets 0 otherwise. 

\paragraph{Dynamics of Two Chains.}
In both hard and reference chains, at any states in $s_{1:d}^{\text{hard}}$ and $s_{1:d}^{\text{ref}}$, all actions $a \notin \mA_{\text{control}}$ invoke transitions to $s_{\text{ter}}$ with 0 rewards. 

In the reference chain, in all environments in $\mG_{\texttt{learn}} \cup \mG_{\texttt{ref}}$, for all actions $a \in \mA_{\text{control}}$, $s_t^{\text{ref}}$ transitions to $s_{t+1}^{\text{ref}}$ with probability $\left(1 - \frac{1}{d + 1 - t}\right)$ and transitions to $s^{\text{ter}}$ otherwise when $t < d$. When the chain transitions to $s_{\text{ter}}$, we receive a reward sampled from $\Ber(1/8)$.

In the hard-to-learn chain, for all environments in $\mG_{\texttt{ref}}$, the system dynamic is identical to the reference chain. The environments in $\mG_{\texttt{learn}}$ are set to be the hard family instances of MDPs from \citet{kwon2021rl} (while setting $d = M/4$), also depicted in Figure~\ref{fig:hard_instance}, Case I:
\begin{enumerate}
    \item At each time, MDPs in $\mG_{\texttt{learn}}$ %\yecomment{each MDP "goes" may be imporoved in terms of writing} 
    transitions from one state in the chain to the next state or to $s_{\text{ter}}$ depending on the played action. When an agent transitions to $s_{\text{ter}}$ it receives a reward drawn from $\Ber(1/8)$. 
    
    \item At all time steps besides at the last one, the agent receives a reward of $0$, when taking an action that does not take it to $s_{\text{ter}}$. At the last time step, if the agent did not move to $s_{\text{ter}}$ and upon taking the action $a_d^*$ it recives a reward of $1$. Hence, the essence of this construction is to identify the optimal action sequence $a_{1:d}^*$ which guarantees a reward $1$ from $\mM_1$ at the end of the chain $s_{d}^{\text{hard}}$. Playing any sub-optimal action sequence generates the distribution of observations indistinguishable from the reference chain. 
\end{enumerate}
We complete the construction in Appendix \ref{appendix:proof:lower_bound}.

\paragraph{Prospective Side Information.} The prospective side information either is a strong prior of one of the MDPs in $\mG_{\texttt{obs}}$, or uninformative in which case $\iota = \iota_{\text{hard}}$. Our construction ensures that when observing $\iota_{\text{hard}}$, all MDPs in $\mG_{\text{learn}}$ and $\mG_{\text{ref}}$ have equal conditional probability, {\it i.e.,} $\PP(m | \iota) = 2/M$ for all $m \in [M/2]$, whereas for other values of prospective information $\iota \neq \iota_{\text{hard}}$, there is one MDP from $\mG_{\text{obs}}$ whose prior probability is greater than $1/2$, and priors over $\mG_{\text{learn}} \cup \mG_{\text{ref}}$ are nearly equally distributed but perturbed by a small parameter $\alpha$, {\it i.e.,} $\PP(m_{\text{obs}} | \iota) \ge 1/2$ for some $m_{\text{obs}} \in [M/2+1, M]$, and $\PP(m | \iota) = O(1/M) + O(\alpha)$ for all $m \in [M/2]$.

\paragraph{Hard Instances.} The family of hard instances $\Theta_{\text{hard}}$ that consists the set of hard-to-learn LMDP-$\Psi$s  is described as follows. All instances in the hard instance family shares the same state space, action space and prospective side-information. The family of hard-to-learn LMDP-$\Psi$s differ in their transition dynamics. Each LMDP-$\Psi$ in $\Theta_{\text{hard}}$ differs by its transition dynamics. The transition dynamics of each element of $\Theta_{\text{hard}}$ is determined by one of the possible sequence  $a^*_{1:d} \in \mA_{\text{control}}^{\bigotimes d}$ that represents the optimal action sequence, and by the `true' exploring actions $a_{\text{explore}}^* \in \mA_{\text{explore}}$. 
% Each instance in $\Theta_{\text{hard}}$ corresponds to the construction presented above with one of possible optimal action sequences.

\paragraph{Reference Model.} We denote $\theta_0$ as the reference model whose hard-to-learn chain is no different from the reference chain in all individual MDPs. In the reference model, at $s_{\text{init}}$, all MDPs in $\mG_{\texttt{ref}}$ transitions to $s_{1}^{\text{hard}}$ and those in $\mG_{\texttt{learn}}$ transitions to $s_{1}^{\text{ref}}$ deterministically when any action in $\mathcal{A}_{\text{explore}}$ is played. All other parts are constructed with the same dynamics as in~$\Theta_{\text{hard}}$.

\paragraph{Proof Overview.} With the above construction, the following lemmas play key roles in proving the regret lower bound:
\begin{lemma}
    \label{lemma:information_equality}
    Let $\psi$ be any exploration strategy for LMDP-$\Psi$. Consider any hard instance $\theta \in \Theta_{\text{hard}}$ and the reference model $\theta_0$. Let $N_{\psi, \iota, a_{1:d}}^{\text{explore}} (K)$ be the number of times that explored the chain systems with the test $t_\iota(a_{1:d}) := \{\iota, a_{\text{explore}}^*, a_{1:d}\}$, {\it i.e.,} with the true exploration action and any sequence $a_{1:d} \in \mA^{\bigotimes d}$ given prospective side information $\iota$. Then,
    \ifarxiv
    \begin{align}
        \sum_{\iota, a_{1:d}} \Exs_{\theta_0} &\left[ N_{\psi, \iota, a_{1:d}}^{\text{explore}} (K) \right] \cdot \KL\left( \PP_{\theta_0} (\cdot | t_\iota(a_{1:d}), \PP_{\theta} (\cdot | t_\iota(a_{1:d}) \right) = \KL \left(\PP^{\psi}_{\theta_0} (\tau^{1:K}), \PP^{\psi}_{\theta} (\tau^{1:K}) \right), \label{eq:data_processing_equality}
    \end{align}
    \else
        \begin{align}
            \sum_{\iota, a_{1:d}} \Exs_{\theta_0} &\left[ N_{\psi, \iota, a_{1:d}}^{\text{explore}} (K) \right] \cdot \KL\left( \PP_{\theta_0} (\cdot | t_\iota(a_{1:d}), \PP_{\theta} (\cdot | t_\iota(a_{1:d}) \right) \nonumber \\
            &= \KL \left(\PP^{\psi}_{\theta_0} (\tau^{1:K}), \PP^{\psi}_{\theta} (\tau^{1:K}) \right), \label{eq:data_processing_equality}
        \end{align}
    \fi
    where $\PP^{\psi} (\tau^{1:K})$ is a distribution of $K$ trajectories obtained with the exploration strategy $\psi$.
\end{lemma}
The main reason for the equality~\eqref{eq:data_processing_equality} is that whenever $a \neq a^{*}_{\text{explore}}$ is played regardless of the prospective side information, the two models $\theta$ and $\theta_0$ generate observations from the same distribution. Then the key lemma is on the bounds for the conditional KL-divergence:
\begin{lemma}
    \label{lemma:kl_divergence_bound}
    For all non optimal action sequences $ a_{1:d} \neq a_{1:d}^*$, the following holds:
    \begin{align*}
        &\KL\left( \PP_{\theta_0} (\cdot | \iota_{\text{hard}}, a_{\text{explore}}^*, a_{1:d} ), \PP_{\theta} (\cdot | \iota_{\text{hard}}, a_{\text{explore}}^*, a_{1:d} ) \right) = 0,
    \end{align*}
    and for all $\iota \in \mI$,
    \begin{align*}
        &\KL\left( \PP_{\theta_0} (\cdot | \iota, a_{\text{explore}}^*, a_{1:d}^* ), \PP_{\theta} (\cdot | \iota, a_{\text{explore}}^*, a_{1:d}^* ) \right) \lesssim \epsilon^2.
    \end{align*}
    Furthermore, for all $\iota\neq \iota_{\text{hard}}$ and $a_{1:d} \neq a_{1:d}^*$:
    \begin{align*}
        \KL\left( \PP_{\theta_0} (\cdot | \iota, a_{\text{explore}}^*, a_{1:d} ), \PP_{\theta} (\cdot | \iota, a_{\text{explore}}^*, a_{1:d} ) \right) \lesssim (\alpha \epsilon)^2.
    \end{align*}
\end{lemma} 
Therefore, we can bound the KL-divergence between the total trajectory distributions of the two models as
\ifarxiv
\begin{align*}
    &\KL \left(\PP^{\psi}_{\theta_0} (\tau^{1:K}), \PP^{\psi}_{\theta} (\tau^{1:K}) \right) \le \Exs_{\theta_0}[N_{\psi, \iota_{\text{hard}}, a_{1:d}^*}^{\text{explore}}] \epsilon^2 + \sum_{\iota \neq \iota_{\text{hard}}, a_{1:d}} \Exs_{\theta_0}[N_{\psi, \iota, a_{1:d}}^{\text{explore}}] (\alpha\epsilon)^2, 
\end{align*}
\else
\begin{align*}
    &\KL \left(\PP^{\psi}_{\theta_0} (\tau^{1:K}), \PP^{\psi}_{\theta} (\tau^{1:K}) \right) \\
    &\le \Exs_{\theta_0}[N_{\psi, \iota_{\text{hard}}, a_{1:d}^*}^{\text{explore}}] \epsilon^2 + \sum_{\iota \neq \iota_{\text{hard}}, a_{1:d}} \Exs_{\theta_0}[N_{\psi, \iota, a_{1:d}}^{\text{explore}}] (\alpha\epsilon)^2, 
\end{align*}
\fi
which translates to the impossibility of distinguishing the two with a probability more than $2/3$ unless either
\ifarxiv
    \begin{align}
         &\Exs_{\theta_0}[N_{\psi, \iota_{\text{hard}}, a_{1:d}^*}^{\text{explore}} (K)] \gtrsim \frac{1}{\epsilon^2}, \text{or } \sum_{\iota \neq \iota_{\text{hard}}, a_{1:d}} \Exs_{\theta_0} [N_{\psi, \iota, a_{1:d}}^{\text{explore}} (K)] \gtrsim \frac{1}{\alpha^2 \epsilon^2}. \label{eq:tv_test_condition}
    \end{align}
\else
\begin{align}
     &\Exs_{\theta_0}[N_{\psi, \iota_{\text{hard}}, a_{1:d}^*}^{\text{explore}} (K)] \gtrsim \frac{1}{\epsilon^2}, \nonumber \\ 
     & \text{or } \sum_{\iota \neq \iota_{\text{hard}}, a_{1:d}} \Exs_{\theta_0} [N_{\psi, \iota, a_{1:d}}^{\text{explore}} (K)] \gtrsim \frac{1}{\alpha^2 \epsilon^2}. \label{eq:tv_test_condition}
\end{align}
\fi
Finally, note that playing sub-optimal actions with $\iota \neq \iota_{\text{hard}}$ incurs at least $1/8$-regret, playing sub-optimal action sequence $a_{1:d} \neq a_{1:d}^*$ incurs at least $O(\epsilon/M)$-regret, and playing the optimal sequence $a_{1:d}^*$ at least $O(1/\epsilon^2)$ times would take $(A^d/\epsilon^2)$ episodes in the worst case. The remaining steps are to formally state the ideas (see Appendix \ref{appendix:proof:lower_bound}).

\section{Conclusion}
% -) Studied LMDPs with prospective information + summary of the results.

In this work, we introduced the LMDP-$\Psi$ setting, when a prospective and weakly revealing information on the latent context is given to an agent. We showed that LMDP-$\Psi$ does not belong to the weakly revealing POMDP class, due to the correlation between observations at different time steps. Further, our results highlight its fundamental different characteristic: we derived an $\Omega(K^{2/3})$ lower bound on its regret, and, hence, the standard $O(\sqrt{K})$ worst-case upper bound is not achievable in general for this class of problems. We also derived a matching $O(K^{2/3})$ upper bound  to complete our results.

From a broader perspective, our results highlight a key deficiency of a ubiquitous assumption made in POMDP modeling, namely, the independence of observation between consecutive time steps, when conditioning on the latent state. We believe that studying the learnability of more general POMDP settings with prospective side information, or non-trivial correlation between observations serves as a fruitful ground for future work. Further, scaling the methods for practical settings, while building on solid theoretical grounds, is a valuable and open research direction.

% \newpage
\bibliographystyle{abbrvnat}
\bibliography{main}

% \newpage
\begin{appendices}
\section{Auxiliary Lemmas}
% \yecomment{make sure there are no references to equations.}

\begin{lemma}[General MLE, \citet{liu2022partially}]
    \label{lemma:mle_traj_concentration}
    With probability $1 - \delta$ for any $\delta > 0$, for all $k \in [K]$, $t \in [H]$ and for any $\theta \in \Theta$, 
    \begin{align}
        \sum_{(\iota, \tau_{t},\pi) \in \mD^k} \log(\PP^\pi_{\theta} (\iota, \tau_{t})) - 3 \log(K|\Theta| / \delta) \le \sum_{(\iota, \tau_{t},\pi) \in \mD^k} \log(\PP^\pi_{\theta^*} (\iota, \tau_{t})).  \label{eq:MLE_concentration}
    \end{align}
\end{lemma}
This is by now a standard MLE technique for constructing confidence sets in RL \cite{agarwal2020flambe}.
\begin{proof}
The proof follows a Chernoff bound type of technique: 
    \begin{align*}
        \PP_{\theta^*} &\left( \sum_{(\iota, \tau_t, \pi) \in \mD^k} \log \left( \frac{\PP^\pi_{\theta} (\iota, \tau_t)}{\PP^\pi_{\theta^*} (\iota, \tau_t)} \right) \ge \Exs_{\theta^*} \left[ \sum_{(\iota, \tau_t,\pi) \in \mD^{k}} \log \left( \frac{\PP^\pi_{\theta} (\iota, \tau_t)}{\PP^\pi_{\theta^*} (\iota, \tau_t)} \right)  \right] + \beta \right) \\
        &\le \PP_{\theta^*} \left( \exp \left( \sum_{(\iota, \tau_t,\pi) \in \mD^k} \log \left( \frac{\PP^\pi_{\theta} (\iota, \tau_t)}{\PP^\pi_{\theta^*} (\iota, \tau_t)} \right) \right) \ge \exp \left( \beta \right) \right) \\
        &\le \Exs_{\theta^*} \left[ \exp \left( \sum_{(\iota, \tau_t,\pi) \in \mD^k} \log \left( \frac{\PP^\pi_{\theta} (\iota, \tau_t)}{\PP^\pi_{\theta^*} (\iota, \tau_t)} \right) \right) \right] \exp(-\beta). 
    \end{align*}
    Note that random variables are $(\iota, \tau_t, \pi)$ in the trajectory dataset $\mathcal{D}^k$, and $$\Exs_{\theta^*} \left[ \sum_{(\tau,\pi) \in \mD^{k}} \log \left( \frac{\PP^\pi_{\theta} (\iota, \tau_t)}{\PP^\pi_{\theta^*} (\iota, \tau_t)} \right)  \right] = - \KL (\PP_{\theta^*} (\mathcal{D}^k) || \PP_{\theta} (\mathcal{D}^k)) \le 0,$$ and further notice the last inequality is by Markov's inequality. Then, 
    \begin{align*}
        \Exs_{\theta^*} \left[ \exp \left( \sum_{(\iota, \tau_t, \pi) \in \mD^k} \log \left( \frac{\PP^\pi_{\theta} (\iota, \tau_t)}{\PP^\pi_{\theta^*} (\iota, \tau_t)} \right) \right) \right] &= \Exs_{\theta^*} \left[ \Pi_{(\iota, \tau_t,\pi) \in \mD^k} \frac{\PP^\pi_{\theta} (\iota, \tau_t)}{\PP^\pi_{\theta^*} (\iota, \tau_t)} \right] = \sum_{\iota, \tau_t} \PP^\pi_{\theta} (\iota, \tau_t) = 1.
    \end{align*}
    Combining the above, taking a union bound over $k \in [K]$ and  $\theta \in \Theta$, letting $\beta = \log (K |\Theta| / \delta)$, with probability $1 - \delta$, the inequality in equality \eqref{eq:MLE_concentration} holds. 
\end{proof}

\begin{lemma}
    \label{lemma:concentration_statistical_distance}
    With probability $1-\delta$, for all $k\in[K]$, $t \in [H]$ and $\theta \in \Theta$, we have
    \begin{align*}
        &\sum_{ \left(\iota, \tau, \pi \right) \in \mathcal{D}^k} \TV^2 \left(\PP_{\theta}^{\pi}(\iota, \tau), \PP_{\theta^*}^{\pi} (\iota, \tau) \right) \lesssim \sum_{(\iota, \tau,\pi) \in \mathcal{D}^k} \log \left( \frac{\PP_{\theta^*}^{\pi}(\iota, \tau)}{\PP_{\theta}^{\pi}(\iota, \tau)} \right) + \beta, \\
        &\sum_{ \left(\iota, \tau, \pi \right) \in \mathcal{D}^k} \HL^2 \left(\PP_{\theta}^{\pi}(\iota, \tau), \PP_{\theta^*}^{\pi} (\iota, \tau) \right) \lesssim \sum_{(\iota, \tau,\pi) \in \mathcal{D}^k} \log \left( \frac{\PP_{\theta^*}^{\pi}(\iota, \tau)}{\PP_{\theta}^{\pi}(\iota, \tau)} \right) + \beta.
    \end{align*}
\end{lemma}
\begin{proof}
    By the TV-distance and Hellinger distance relation, for any $\iota, \tau$, $\pi$ and $t\in[H]$, 
    \begin{align*}
        \TV^2 \left(\PP_{\theta}^{\pi}(\iota, \tau), \PP_{\theta^*}^{\pi} (\iota, \tau) \right) &\le 2 \HL^2 \left(\PP_{\theta}^{\pi}(\iota, \tau), \PP_{\theta^*}^{\pi} (\iota, \tau) \right) \\
        &= 2\left(1 - \Exs_{\iota, \tau \sim \PP_{\theta^*}^{\pi}} \left[ \sqrt{\frac{\PP_{\theta}^{\pi} (\iota, \tau)}{\PP_{\theta^*}^{\pi} (\iota, \tau)}} \right] \right) \\
        &\le -2 \log \left( \Exs_{\iota, \tau \sim \PP_{\theta^*}^{\pi}} \left[ \sqrt{\frac{\PP_{\theta}^{\pi} (\iota, \tau)}{\PP_{\theta^*}^{\pi} (\iota, \tau)}} \right] \right).
    \end{align*}
    To bound the summation over samples, we start from
    \begin{align*}
        \sum_{ \left(\iota, \tau, \pi \right) \in \mathcal{D}^k} \TV^2 \left(\PP_{\theta}^{\pi}(\iota, \tau), \PP_{\theta^*}^{\pi} (\iota, \tau) \right) &\le -2  \sum_{\left(\iota, \tau, \pi \right) \in \mathcal{D}^k} \log \left( \Exs_{\iota, \tau \sim \PP_{\theta^*}^\pi} \left[ \sqrt{\frac{\PP_{\theta}^\pi (\iota, \tau)}{\PP_{\theta^*}^\pi (\iota, \tau)}} \right] \right). 
    \end{align*}
    On the other hand, by the Chernoff bound, 
    \begin{align*}
        \PP_{\theta^*} &\left( \sum_{(\iota, \tau,\pi) \in \mD^k} \log \left( \sqrt{\frac{\PP_{\theta}^\pi (\iota, \tau)}{\PP_{\theta^*}^\pi (\iota, \tau)}} \right) \ge \sum_{(\iota, \tau,\pi) \in \mD^{k}} \log \Exs_{\iota, \tau \sim \PP_{\theta^*}^\pi } \left[ \sqrt{\frac{\PP_{\theta}^\pi (\iota, \tau)}{\PP_{\theta^*}^\pi (\iota, \tau)}} \right] + \beta \right) \\
        &\le \Exs_{\theta^*} \left[ \frac{\exp \left( \sum_{(\iota, \tau, \pi) \in \mD^k} \log \left( \sqrt{ \frac{\PP^\pi_{\theta} (\iota, \tau)}{\PP^\pi_{\theta^*} (\iota, \tau)} } \right) \right)}{\exp \left( \sum_{(\iota, \tau,\pi) \in \mD^{k}} \log \Exs_{ \iota, \tau \sim \PP_{\theta^*}^\pi } \left[\sqrt{\frac{\PP_{\theta}^\pi ( \iota, \tau)}{\PP_{\theta^*}^\pi ( \iota, \tau)}}   \right] \right)} \right] \exp(-\beta) \\
        &= \Exs_{\theta^*} \left[ \frac{\Pi_{(\iota, \tau,\pi) \in \mD^k} \sqrt{\frac{\PP_{\theta}^\pi ( \iota, \tau)}{\PP_{\theta^*}^\pi ( \iota, \tau)}} }{ \Pi_{(\iota, \tau,\pi) \in \mD^{k}} \Exs_{\iota, \tau \sim \PP_{\theta^*}^\pi } \left[\sqrt{\frac{\PP_{\theta}^\pi (\iota, \tau)}{\PP_{\theta^*}^\pi (\iota, \tau)}}  \right] } \right] \exp(-\beta) \\
        &= \Exs_{\theta^*} \left[ \frac{ \Pi_{(\iota, \tau,\pi) \in \mD^{k-1}} \sqrt{\frac{\PP_{\theta}^\pi (\iota, \tau)}{\PP_{\theta^*}^\pi ( \iota, \tau)}}  \cdot \Exs_{\iota, \tau^k \sim \PP_{\theta^*}^{\pi^k} } \left[ \sqrt{ \frac{\PP^{\pi^k}_{\theta} (\iota, \tau^k)}{\PP^\pi_{\theta^*} (\iota, \tau^k)} } \Big| \pi^k, \mathcal{\mD}^{k-1} \right] }{ \Pi_{(\iota, \tau,\pi) \in \mD^{k}} \Exs_{\iota, \tau \sim \PP_{\theta^*}^\pi } \left[\sqrt{\frac{\PP^\pi_{\theta} (\iota, \tau)}{\PP^\pi_{\theta^*} (\iota, \tau )}}  \right] } \right] \exp(-\beta) \\
        &= \Exs_{\theta^*} \left[ \frac{\Pi_{(\iota, \tau, \pi) \in \mD^{k-1}} \sqrt{ \frac{\PP^\pi_{\theta} (\iota, \tau)}{\PP^\pi_{\theta^*} (\iota, \tau)} } }{ \Pi_{(\iota, \tau, \pi) \in \mD^{k-1}} \Exs_{\iota, \tau \sim \PP_{\theta^*}^\pi } \left[\sqrt{\frac{\PP^\pi_{\theta} (\iota, \tau)}{\PP^\pi_{\theta^*}(\iota, \tau)}}  \right] } \right] \exp(-\beta) = ... = \exp(-\beta),
    \end{align*}
    where in the last line, we used the tower property of expectation. Thus, again by setting $\beta = O\left(\log (KH|\Theta| / \delta) \right)$, with probability at least $1 - \delta$, we have
    \begin{align*}
        \sum_{(\iota, \tau, \pi) \in \mathcal{D}^k} & \TV^2 (\PP_{\theta}^\pi (\iota, \tau ), \PP_{\theta^*}^\pi (\iota, \tau)) \lesssim -\frac{1}{2} \sum_{(\iota, \tau,\pi) \in \mD^k} \log \left( \frac{\PP^\pi_{\theta} (\iota, \tau)}{\PP^\pi_{\theta^*} (\iota, \tau)} \right) + \beta \\
        &= -\frac{1}{2} \sum_{(\iota, \tau ,\pi) \in \mD^k} \log \left( \frac{\PP^\pi_{\theta} (\iota, \tau)}{\PP^\pi_{\theta^*} (\iota, \tau)} \right) + \frac{1}{2} \sum_{(\iota, \tau,\pi) \in \mD^k} \log \left( \frac{\PP^\pi_{\theta} (\iota, \tau)}{\PP^\pi_{\theta^*} (\iota, \tau)} \right) + \beta.
    \end{align*}
    We can apply Lemma \ref{lemma:mle_traj_concentration}, and finally have
    \begin{align*}
        \sum_{(\iota, \tau, \pi) \in \mathcal{D}^k} & \TV^2 (\PP_{\theta}^\pi (\iota, \tau), \PP_{\theta^*}^\pi (\iota, \tau)) \lesssim -\sum_{(\iota, \tau,\pi) \in \mD^k} \log \left( \frac{\PP^\pi_{\theta} (\iota, \tau)}{\PP^\pi_{\theta^*} (\iota, \tau)} \right) + \beta.
    \end{align*}
\end{proof}

Most of the following lemmas can also be found in \cite{huang2023provably} as we adopt their proof strategy. We state and prove them for the completeness. The following is the concentration lemma for the empirical {\it conditional} probability, which Importantly, this property still holds regardless of causal relationships inside each trajectory sample:
\begin{lemma}
    \label{lemma:concentration_conditioanl_tv_dist}
    With probability $1-\delta$, for all $k\in[K]$, $t \in [H]$, $\theta \in \Theta$, we have
    \begin{align*}
        &\sum_{ \left(\iota, \tau_t, \omega_t, \pi \right) \in \mathcal{D}^k} \TV^2 \left(\PP_{\theta}^{\pi}(\iota, \omega_t | \tau_t), \PP_{\theta^*}^{\pi} (\iota, \omega_t | \tau_t) \right) \lesssim \sum_{(\iota, \tau,\pi) \in \mathcal{D}^k} \log \left( \frac{\PP_{\theta^*}^{\pi}(\iota, \tau)}{\PP_{\theta}^{\pi}(\iota, \tau)} \right) + \beta, \\
        &\sum_{ \left(\iota, \tau_t, \omega_t, \pi \right) \in \mathcal{D}^k} \HL^2 \left(\PP_{\theta}^{\pi}(\iota, \omega_t | \tau_t), \PP_{\theta^*}^{\pi} (\iota, \omega_t | \tau_t) \right) \lesssim \sum_{(\iota, \tau,\pi) \in \mathcal{D}^k} \log \left( \frac{\PP_{\theta^*}^{\pi}(\iota, \tau)}{\PP_{\theta}^{\pi}(\iota, \tau)} \right) + \beta.
    \end{align*}
\end{lemma}
\begin{proof}
    The proof is almost identical except that we now start from
    \begin{align*}
        \sum_{ \left(\tau, \pi \right) \in \mathcal{D}^k} \TV^2 \left(\PP_{\theta}^{\pi}(\iota,\omega_t | \tau_t), \PP_{\theta^*}^{\pi} (\iota,\omega_t | \tau_t) \right) &\le -2  \sum_{\left(\tau, \pi \right) \in \mathcal{D}^k} \log \left( \Exs_{(\iota,\omega_t) \sim \PP_{\theta^*}^\pi(\cdot | \tau_t)} \left[ \sqrt{\frac{\PP_{\theta}^\pi (\iota, \omega_t | \tau_t)}{\PP_{\theta^*}^\pi (\iota, \omega_t | \tau_t)}} \right] \right). 
    \end{align*}
    and use the tower property of expectation conditioned on $\tau_t^k$. Thus, again by setting $\beta = O\left(\log (KH|\Theta| / \delta) \right)$, with probability at least $1 - \delta$, we have
    \begin{align*}
        &\sum_{ \left(\tau, \pi \right) \in \mathcal{D}^k} \TV^2 \left(\PP_{\theta}^{\pi}(\iota,\omega_t | \tau_t), \PP_{\theta^*}^{\pi} (\iota,\omega_t | \tau_t) \right) \lesssim -\frac{1}{2} \sum_{(\tau,\pi) \in \mD^k} \log \left( \frac{\PP^\pi_{\theta} (\iota,\omega_t | \tau_t)}{\PP^\pi_{\theta^*} (\iota,\omega_t | \tau_t)} \right) + \beta \\
        &= -\frac{1}{2} \sum_{(\iota, \tau ,\pi) \in \mD^k} \log \left( \frac{\PP^\pi_{\theta} (\iota, \tau)}{\PP^\pi_{\theta^*} (\iota, \tau)} \right) + \frac{1}{2} \sum_{(\iota, \tau,\pi) \in \mD^k} \log \left( \frac{\PP^\pi_{\theta} (\tau_t)}{\PP^\pi_{\theta^*} (\tau_t)} \right) + \beta.
    \end{align*}
    Finally, we apply Lemma \ref{lemma:mle_traj_concentration}, and have
    \begin{align*}
        \sum_{ \left(\iota,\tau, \pi \right) \in \mathcal{D}^k} \TV^2 \left(\PP_{\theta}^{\pi}(\iota,\omega_t | \tau_t), \PP_{\theta^*}^{\pi} (\iota,\omega_t | \tau_t) \right) \lesssim -\sum_{(\iota,\tau,\pi) \in \mD^k} \log \left( \frac{\PP^\pi_{\theta} (\iota,\tau)}{\PP^\pi_{\theta^*} (\iota,\tau)} \right) + \beta.
    \end{align*}
\end{proof}

\begin{lemma}
    \label{lemma:loose_HL_conditional_to_total_bound}
    For arbitrary probability distribution $P, Q$ over joint distributions $(\tau, \omega)$, 
    \begin{align*}
        \Exs_{\tau\sim P} [ \HL^2 (P(\omega|\tau), Q(\omega|\tau)) ] \le 4 \HL^2 (P(\omega, \tau), Q(\omega, \tau)).
    \end{align*}
\end{lemma}
\begin{proof}
    We prove this statement by explicitly bounding the Hellinger distance.
    \begin{align*}
        &\int \left(\int \left( \sqrt{P(\omega|\tau)} - \sqrt{Q(\omega | \tau)} \right)^2 d\omega \right) P(\tau) d\tau \\
        &\le 2\int \int \left( \sqrt{P(\omega, \tau)} - \sqrt{Q(\tau) Q(\omega | \tau)} \right)^2  d\omega d\tau
        +  2\int \int \left( \sqrt{P(\tau) Q(\omega| \tau)} - \sqrt{Q(\tau) Q(\omega | \tau)} \right)^2 d\omega d\tau \\
        &= 2\HL^2(P(\omega,\tau), Q(\omega,\tau))
        +  2\int \int \left( \sqrt{P(\tau)} - \sqrt{Q(\tau)} \right)^2 Q(\omega | \tau) d\omega d\tau \\
        &\le 4\HL^2(P(\omega,\tau), Q(\omega,\tau)).
    \end{align*}
\end{proof}

\begin{lemma}
    \label{lemma:elliptical_potential_expectation}
    Let $x \in \mathbb{R}^d$ be a random vector from a series of distributions $\{\mD^k\}_k$ and let $U_k = U_1 + \sum_{j < k} \Exs_{x \sim \mD^j}[x x^\top]$ with $U_1 \succeq \lambda I$ for some positive constant $\lambda > 0$. Assume that $\|x\|_2 \le 1$ almost surely. Then,
    \begin{align*}
        \sum_{k=1}^K \min\left( \Exs_{x \sim \mD^k} \left[\|x\|^2_{U_k^{-1}} \right], R\right) \le (1+R) d \log(1 + K/\lambda). 
    \end{align*}
\end{lemma}
This is minor variation of the standard result from \cite{abbasi2011improved}. Differently from their result, here, we need to establish the bound for the expected $U_k$. Hence their result is not directly applied here.
\begin{proof}
     We follow the same technique of \cite{abbasi2011improved}.
    \begin{align*}
        \sum_{k=1}^K \min\left( \Exs_{x \sim \mD^k} \left[\|x\|^2_{U_k^{-1}} \right], R\right) &\le (1+R) \sum_{k=1}^K \log \left(1 + \Exs_{x \sim \mD^k} \left[\|x\|^2_{U_k^{-1}} \right] \right) \\
        &\stackrel{(a)}{=} (1+R) \sum_{k=1}^K \log \left(1 +  \texttt{Tr} ( \Exs_{x \sim \mD^k} \left[x x^\top \right] U_k^{-1}) \right) \\
        &= (1+R) \sum_{k=1}^K \log \left(1 +  \texttt{Tr} ( (U_{k+1} - U_k) U_k^{-1}) \right) \\
        &\le (1+R) \sum_{k=1}^K \log \texttt{det} \left(I_d +  ( U_k^{-1/2} (U_{k+1} - U_k) U_k^{-1/2}) \right) \\
        &= (1+R) \sum_{k=1}^K \log \frac{\texttt{det} U_{k+1}}{\texttt{det} (U_k)} = (1+R) \log \frac{\texttt{det} (U_{K+1})}{\texttt{det} (U_1)} \\
        &\le (1+R) d \log(1 + K / \lambda),
    \end{align*}
    where $(a)$ is due to the linearity of trace operators. 
\end{proof}

\begin{lemma}
    \label{lemma:trace_sum_bound}
    Let $x_k$ be any sequence of vectors in $\mathbb{R}^d$ where $\textbf{rank}(\{x_k\}_k) = r < d$, and let $U_k = \lambda I + \sum_{j < k} x_j x_j^\top$. Then,
    \begin{align*}
        \sum_{j<k} \|x_k\|_{U_k^{-1}}^2 \le r.
    \end{align*}
\end{lemma}
\begin{proof}
    Again, we can express $a^\top A^{-1} a = \texttt{Tr}(aa^\top A^{-1})$, and thus
    \begin{align*}
        \sum_{j<k} \texttt{Tr}(x_jx_j^\top U_k^{-1}) &= \texttt{Tr}\left((\tssum_{j<k} x_jx_j^\top) U_k^{-1} \right) \\
        &= \texttt{Tr} \left(I - \left(I + \lambda^{-1} \tssum_{j<k} x_jx_j^\top\right)^{-1} \right) \le r,
    \end{align*}
    where the inequality holds since the matrix inside $\texttt{Tr}$ is at most rank $r$ with eigenvalues less than or equal to one.
    %\yecomment{add few lines to justify the upper bound}
\end{proof}

\begin{lemma}
    \label{lemma:bonus_comparison_lemma}
    For any vectors $a,b$ and positive definite matrices $A,B$ such that $A, B \succeq \lambda_0 I$, we have
    \begin{align*}
        \|a\|_{A^{-1}} - \|b\|_{B^{-1}} \le \frac{1}{\sqrt{\lambda_0}} \|a - b\|_2 + \|b\|_{B^{-1}} \|A^{-1/2} (B-A) B^{-1/2}\|_2.
    \end{align*}
\end{lemma}
\begin{proof}
    The proof follows by algebraic manipulations:
    \begin{align*}
        \|a\|_{A^{-1}} - \|b\|_{B^{-1}} &= \frac{\|a\|_{A^{-1}}^2 - \|b\|_{B^{-1}}^2 }{\|a\|_{A^{-1}} + \|b\|_{B^{-1}}} \\
        &= \frac{a^\top A^{-1} (a-b) + (a-b)^\top B^{-1} b + a^\top A^{-1} (B - A) B^{-1} b}{ \|a\|_{A^{-1}} + \|b\|_{B^{-1}}} \\
        &\le \frac{\|a\|_{A^{-1}} \|a-b\|_{A^{-1}} + \|a-b\|_{B^{-1}} \|b\|_{B^{-1}} + a^\top A^{-1} (B - A) B^{-1} b}{\|a\|_{A^{-1}} + \|b\|_{B^{-1}}} \\
        &\le \frac{1}{\sqrt{\lambda_0}} \|a - b\|_2 + \|b\|_{B^{-1}} \|A^{-1/2} (B-A) B^{-1/2}\|_2. 
    \end{align*}
\end{proof}

\section{Proof of Upper Bounds}
\label{appendix:proof:upper_bound}
We remind the reader some notations we frequently use in the appendix.
\begin{align*}
    &B(o,s_{+1}|s,a) = \mathbb{I} \cdot \diag ([\PP(o,s_{+1} | m,s,a)]_{m=1}^M) \cdot \mathbb{I}^{\dagger}, \\
    &b_0 = \mathbb{I} w, \\
    &\tau_t = (s_1, a_1, o_1, ..., s_t, a_t), \\
    &\omega_t = (o_{t}, s_{t+1}, a_{t+1}, ..., o_H), \\
    &\psi (\omega_t, \iota | s_t)^\top = \bm{e}_{\iota}^\top \cdot \Pi_{h=t}^{H} B {(o_{h},s_{h+1}|s_{h},a_{h})}, \\
    &b(\tau_{t}) = \Pi_{h=1}^{t-1} B {(o_{h}, s_{h+1}|s_{h},a_{h})} b_0, \\
    &\pi(\tau_t) = \Pi_{h=1}^{t} \pi(a_{h} | s_1, ..., s_{h}), \\
    &\pi(\omega_t | \tau_t) = \Pi_{h=t+1}^{H} \pi(a_{h} | s_1, ..., s_{h}).
\end{align*}
We frequently use a shorthand for a pair of observations, $x_t := (s_t, a_t)$ and $y_t := (o_t, s_{t+1})$.

\subsection{Proof of Theorem \ref{theorem:regret_upper_bound_blind}}
There are several analysis techniques available in previous work ({\it e.g.,} \citet{liu2022partially, uehara2022provably, liu2023optimistic, chen2022partially, huang2023provably}). Among all the above great works, we find the recent analysis of \citet{huang2023provably} as particularly well-suited for our setting, and thus we adopt their proof ideas. 

By the choice of model selection in the confidence set, it is sufficient to bound the sum TV-distances since
\begin{align*}
    \sum_{k=1}^K V_{\theta^*}^{\pi_{\texttt{blind}}^*} - V_{\theta^*}^{\pi^k} &\le \sum_{k=1}^K V_{\theta^k}^{\pi_{k}} - V_{\theta^*}^{\pi^k} \le H \cdot \sum_{k=1}^K \TV(\PP_{\theta^k}^{\pi^k}, \PP_{\theta^*}^{\pi^k}).
\end{align*}
At each episode $k \in [K]$, we start by unfolding the upper bound of the total-variation distance:
\begin{align*}
    \TV(\PP_{\theta^*}^{\pi^k}(\tau, \iota), \PP_{\theta}^{\pi^k}(\tau, \iota)) &\le \sum_{\tau, \iota} \sum_{t=1}^H  \pi(\tau) \cdot \left| \psi_{\theta^k}(\omega_{t+1}, \iota| x_{t+1})^\top b_{\theta^*}(\tau_{t+1}) - \psi_{\theta^k}(\omega_{t}, \iota| x_{t})^\top b_{\theta^*}(\tau_{t}) \right|  \\
    &= \sum_{t=1}^H \sum_{\tau, \iota} \pi(\tau) \cdot \left| \psi_{\theta^k}(\omega_{t+1}, \iota| x_{t+1})^\top \left(B_{\theta^k} (y_t|x_t) - B_{\theta^*} (y_t|x_t) \right) b_{\theta^*}(\tau_{t}) \right|.
\end{align*}
We focus on bounding the inner summation fixing $t$. Every trajectory $\tau$ can be decomposed into $\tau_t$ and $\omega_t$, and thus
\begin{align}
    &\TV(\PP_{\theta^*}^{\pi^k}(\tau, \iota), \PP_{\theta}^{\pi^k}(\tau, \iota)) \nonumber \\
    &\le \sum_{t} \sum_{\tau_t} \pi^k(\tau_t) \sum_{\omega_{t},\iota} \pi^k(\omega_t | \tau_t) \cdot \left| \psi_{\theta^k}(\omega_{t+1}, \iota | x_{t+1})^\top \left(B_{\theta^k} (y_t|x_t) - B_{\theta^*} (y_t|x_t) \right) \mathbb{I}_{\theta^*} \mathbb{I}_{\theta^*}^{\dagger} b_{\theta^*}(\tau_{t}) \right|, \label{eq:tv_to_operator_diff_blind}
\end{align}
where we used $\mathbb{I}_{\theta^*} \mathbb{I}_{\theta^*}^{\dagger} b_{\theta^*}(\cdot) = b_{\theta^*}(\cdot)$ since $b_{\theta^*}(\cdot)$ is in the column span of $\mathbb{I}_{\theta^*}$. Define
\begin{align*}
    v_{\theta^*}(\tau_t) = \mathbb{I}_{\theta^*}^{\dagger} b_{\theta^*}(\tau_t), \text{ and }, \bar{v}_{\theta^*}(\tau_t) = \frac{v_{\theta^*}(\tau_t)}{\|v_{\theta^*}(\tau_t)\|_1},
\end{align*}
which are the internal unnormalized and normalized latent belief states, respectively. Then the RHS in equation \eqref{eq:tv_to_operator_diff_blind} can be expressed as
\begin{align*}
    \sum_{\tau_t} \pi^k(\tau_t) \|\bar{v}_{\theta^*}(\tau_t)\|_1 \sum_{\omega_t, \iota} \pi^k(\omega_t | \tau_t) \cdot \left| \psi_{\theta^k} (\omega_{t+1}, \iota | x_{t+1})^\top \left(B_{\theta^k} (y_t|x_t) - B_{\theta^*} (y_t|x_t) \right) \mathbb{I}_{\theta^*} \bar{v}_{\theta^*}(\tau_{t}) \right|.
\end{align*}
Define an elliptical potential matrix $\Lambda_*^k(s,a)$ as
\begin{align*}
    \Lambda_*^k(s,a) = \lambda^* I + \sum_{j < k} \Exs_{\theta^*}^{\pi^j} \left[ \indic{(s_t,a_t) = (s,a)} \bar{v}_{\theta^*}(\tau_{t}) \bar{v}_{\theta^*}(\tau_{t})^\top \right],
\end{align*}
where we define $\lambda^*$ later (here, the choice of $\lambda^*$ does not matter much). Using Cauchy-Schwartz inequality, we can separate the concentration argument and the pigeon-hole (a.k.a. elliptical potential lemma) argument. For simplicity, let $f(\omega_{t}, \iota) := \psi_{\theta^k} (\omega_{t+1}, \iota | x_{t+1})^\top \left(B_{\theta^k} (y_t|x_t) - B_{\theta^*} (y_t|x_t) \right) \mathbb{I}_{\theta^*}$. Then
\begin{align*}
    &\sum_{\omega_t, \iota} \pi^k(\omega_t | \tau_t) \cdot \left| \psi_{\theta^k} (\omega_{t+1}, \iota | x_{t+1})^\top \left(B_{\theta^k} (y_t|x_t) - B_{\theta^*} (y_t|x_t) \right) \mathbb{I}_{\theta^*} \bar{v}_{\theta^*}(\tau_{t}) \right| \\
    &= \sum_{\omega_t, \iota} \pi^k(\omega_t | \tau_t) \cdot \left| f(\omega_t,\iota) \bar{v}_{\theta^*}(\tau_{t}) \right| = \sum_{\omega_t, \iota} \pi^k(\omega_t | \tau_t) \cdot f(\omega_t,\iota) \sign(f(\omega_t,\iota) \bar{v}_{\theta^*}(\tau_{t})) \cdot \bar{v}_{\theta^*}(\tau_{t}) \\
    &\le \left\|\sum_{\omega_t, \iota} \pi^k(\omega_t | \tau_t) \cdot f(\omega_t,\iota) \sign(f(\omega_t,\iota) \bar{v}_{\theta^*}(\tau_{t})) \right\|_{\Lambda_*^k(x_t)} \left\|\bar{v}_{\theta^*}(\tau_{t}) \right\|_{\Lambda_*^k(x_t)^{-1}}.
\end{align*}
Checking the squared norm of the first part, we observe that
\begin{align*}
    &\left\|\sum_{\omega_t, \iota} \pi^k(\omega_t | \tau_t) \cdot f(\omega_t,\iota) \sign(f(\omega_t,\iota) \bar{v}_{\theta^*}(\tau_{t})) \right\|_{\Lambda_*^k(x_t)}^2 \\
    &= \underbrace{\lambda^* \left\|\sum_{\omega_t, \iota} \pi(\omega_t|\tau_t) f(\omega_t,\iota) \cdot \sign(f(\omega_t,\iota) \bar{v}_{\theta^*}(\tau_t)) \right\|_2^2}_{(i)} \\
    &\quad + \underbrace{\sum_{j<k} \Exs_{\theta^*}^{\pi^j} \left[\indic{x_t^j = x_t} \left( \sum_{\omega_t, \iota} \pi(\omega_t|\tau_t) (f(\omega_t,\iota) \bar{v}_{\theta^*}(\tau_t^j)) \cdot \sign(f(\omega_t,\iota) \bar{v}_{\theta^*}(\tau_t)) \right)^2\right]}_{(ii)}.
\end{align*}

\paragraph{Bounding $(i)$.} For any $m \in [M]$ we observe that
\begin{align*}
    &\left| \sum_{\omega_t, \iota} \pi(\omega_t|\tau_t) f(\omega_t,\iota) \bm{e}_m \cdot \sign(f(\omega_t,\iota) \bar{v}_{\theta^*}(\tau_t)) \right| \le \sum_{\omega_t, \iota} \left| \pi(\omega_t|\tau_t) f(\omega_t,\iota) \bm{e}_m \right| \\
    &\le \sum_{\omega_t,\iota} \pi(\omega_t | \tau_t) \left| \psi_{\theta^k}(\omega_{t+1}, \iota | x_{t+1})^\top \left(B_{\theta^k} (y_t| x_t) - B_{\theta^*} (y_t| x_t) \right) \mathbb{I}_{\theta^*} \bm{e}_{m}) \right| \\
    &\le \sum_{\omega_t, \iota} \pi(\omega_t | \tau_t) \left| \psi_{\theta^k}(\omega_{t}, \iota | x_{t})^\top \mathbb{I}_{\theta^*} \bm{e}_{m} - \psi_{\theta^k}(\omega_{t+1}, \iota | x_{t+1})^\top \mathbb{I}_{\theta^*} \bm{e}_m \cdot \PP_{\theta^*} (y_t|m,x_t) \right| \\
    &\le \frac{2M}{\alpha} \|\mathbb{I}_{\theta} \bm{e}_m\|_1 = \frac{2M}{\alpha}.
\end{align*}
Therefore, $(i) \le \lambda^* M (2M/\alpha)^2 = 4M^3 \lambda^* / \alpha^2$. 

\paragraph{Bounding $(ii)$.} Observe that
\begin{align*}
    &\sum_{\omega_t, \iota} \pi(\omega_t|\tau_t) (f(\omega_t,\iota) \bar{v}_{\theta^*}(\tau_t^j)) \cdot \sign(f(\omega_t,\iota) \bar{v}_{\theta^*}(\tau_t)) \\
    &\le \sum_{\omega_t, \iota} \pi(\omega_t|\tau_t) \left| \psi_{\theta^k}(\omega_{t+1}, \iota | x_{t+1})^\top \left( B_{\theta^k} (y_t | x_t) - B_{\theta^*} (y_t| x_t) \right) \mathbb{I}_{\theta^*} \bar{v}_{\theta^*} (\tau_t^j) \right| \\
    &\le \sum_{\omega_t, \iota} \pi(\omega_t|\tau_t) \left| \psi_{\theta^k}(\omega_{t+1}, \iota | x_{t+1})^\top \left( B_{\theta^k} (y_t | x_t) \bar{b}_{\theta^k} (\tau_t^j) - B_{\theta^*} (y_t| x_t) \bar{b}_{\theta^*} (\tau_t^j) \right) \right| \\
    &\quad + \sum_{\omega_t, \iota} \pi(\omega_t|\tau_t) \left| \psi_{\theta^k}(\omega_{t}, \iota | x_{t})^\top \left( \bar{b}_{\theta^k} (\tau_t^j) - \bar{b}_{\theta^*} (\tau_t^j) \right) \right| \\
    &\le \frac{M}{\alpha} \left(\|\bar{b}_{\theta^k} (\tau_t^j) - \bar{b}_{\theta^*} (\tau_t^j)\|_1  +  \sum_{y_t} \|B_{\theta^k} (y_t | x_t) \bar{b}_{\theta^k} (\tau_t^j) - B_{\theta^*} (y_t| x_t) \bar{b}_{\theta^*} (\tau_t^j)\|_1  \right),
\end{align*}
where we denoted $\bar{b}_{\theta} = \mathbb{I}_{\theta} \bar{v}_{\theta}$ for any $\theta$. The last inequality follows from the well-conditionedness of the system following equation \eqref{eq:well_conditioned_PSR}. Then the statistical meaning of each term is given by
\begin{align*}
    &\bm{e}_{\iota}^\top \bar{b}_{\theta} (\tau_t^j) = \PP_{\theta}^{\pi^j} (\iota | \tau_t^j), \\
    &\indic{x_t^j = x_t} \bm{e}_{\iota}^\top B_{\theta}(y_t | x_t) \bar{b}_{\theta} (\tau_t^j) = \PP_{\theta}^{\pi_j}(\iota, y_t | \tau_t^j). 
\end{align*}
The second equality can be verified by the following steps:
\begin{align*}
    &\indic{x_t^j = x_t} \cdot \bm{e}_{\iota}^\top B_{\theta} (y_t |x_t) \bar{b}_{\theta}(\tau_t^j) \\
    &= \frac{\mathbf{1}^\top \diag(\PP_{\theta} ( \iota | m)) \Pi_{h=1}^{t} \diag(\PP_{\theta} (y_{h} |m,x_{h}^j )) w }{ \| \bar{b}_{\theta}(\tau_t^j) \|_1 } \\
    &= \frac{\mathbf{1}^\top \diag(\PP_{\theta} ( \iota | m)) \Pi_{h=1}^{t} \diag(\PP_{\theta} (y_{h} |m,x_{h}^j )) w}{\sum_{\iota'} \mathbf{1}^\top \diag(\PP_{\theta} (\iota'|m)) \Pi_{h=1}^{t-1} \diag_{\theta} (\PP(y_{h} |m,x_{h}^j)) w} \\
    &= \frac{\pi^j(\tau_t^j) \mathbf{1}^\top \diag(\PP_{\theta} ( \iota | m)) \Pi_{h=1}^{t} \diag(\PP_{\theta} (y_{h} |m,x_{h}^j )) w}{\sum_{\iota'} \pi^j(\tau_t^j) \mathbf{1}^\top \diag(\PP_{\theta} (\iota'|m)) \Pi_{h=1}^{t-1} \diag_{\theta} (\PP(y_{h} |m,x_{h}^j)) w} \\
    &= \frac{\PP^{\pi^j}_{\theta} (\iota, y_{t}, \tau_t^j)}{\PP^{\pi^j}_{\theta} (\tau_t^j)} = \PP^{\pi^j}_{\theta} (i, y_t | \tau_t^j).
\end{align*}
In summary, we have $(ii) \le \frac{4M^2}{\alpha^2} \left( \TV^2 (\PP_{\theta^k}^{\pi^j} (\iota, y_t | \tau_t^j), \PP_{\theta^*}^{\pi^j} (\iota, y_t | \tau_t^j)) \right)$.

\paragraph{Combining bounds for (i) and (ii).} Therefore, we can conclude that
\begin{align*}
    &\left\|\sum_{\omega_t, \iota} \pi^k(\omega_t | \tau_t) \cdot f(\omega_t,\iota) \sign(f(\omega_t,\iota) \bar{v}_{\theta^*}(\tau_{t})) \right\|_{\Lambda_*^k(x_t)}^2 \\
    &\le \frac{4M^3 \lambda^*}{\alpha^2} + \frac{4M^2}{\alpha^2} \sum_{j < k} \Exs_{\theta^*}^{\pi^j} \left[ \TV^2 (\PP_{\theta^k}^{\pi^j} (\iota, y_t | \tau_t^j), \PP_{\theta^*}^{\pi^j} (\iota, y_t | \tau_t^j)) \right] \\
    &\le \frac{4M^3 \lambda^*}{\alpha^2} + \frac{8M^2}{\alpha^2} \sum_{j < k} \Exs_{\theta^*}^{\pi^j} \left[ \HL^2 (\PP_{\theta^k}^{\pi^j} (\iota, y_t | \tau_t^j), \PP_{\theta^*}^{\pi^j} (\iota, y_t | \tau_t^j)) \right] \\
    &\le \frac{4M^3 \lambda^*}{\alpha^2} + \frac{32M^2}{\alpha^2} \sum_{j < k} \HL^2 (\PP_{\theta^k}^{\pi^j} (\iota, y_t, \tau_t^j), \PP_{\theta^*}^{\pi^j} (\iota, y_t, \tau_t^j)),
\end{align*}
where we used Lemma \ref{lemma:loose_HL_conditional_to_total_bound}. Finally, due to the concentration of the square sum of Helligner distances (Lemma \ref{lemma:concentration_statistical_distance}), we can conclude that
\begin{align*}
    \left\|\sum_{\omega_t, \iota} \pi^k(\omega_t | \tau_t) \cdot f(\omega_t,\iota) \sign(f(\omega_t,\iota) \bar{v}_{\theta^*}(\tau_{t})) \right\|_{\Lambda_*^k(x_t)}^2 \lesssim \frac{M^2}{\alpha^2} (\lambda^* M + \beta). 
\end{align*}
Plugging this bound back to equation~\eqref{eq:tv_to_operator_diff_blind}, we have
\begin{align*}
    \TV(\PP_{\theta^*}^{\pi^k}(\tau, \iota), \PP_{\theta^k}^{\pi^k}(\tau, \iota)) &\lesssim \frac{M}{\alpha} \sqrt{(\lambda^* M + \beta)}\cdot \sum_t \sum_{\tau_t} \pi^k(\tau_t) \| \bar{v}_{\theta^*}(\tau_t) \|_1 \|\bar{v}_{\theta^*}(\tau_t)\|_{\Lambda_*^k(x_t)^{-1}} \\
    &= \frac{M}{\alpha} \sqrt{(\lambda^* M + \beta)}\cdot \sum_t \Exs_{\theta^*}^{\pi^k} \left[ \|\bar{v}_{\theta^*}(\tau_t)\|_{\Lambda_*^k(x_t)^{-1}} \right]. 
\end{align*}
Finally, summing up over all episodes, we have
\begin{align*}
    \sum_{k=1}^K \TV(\PP_{\theta^*}^{\pi^k}(\tau, \iota), \PP_{\theta^k}^{\pi^k}(\tau, \iota)) &\lesssim \frac{M}{\alpha} \sqrt{(\lambda^* M + \beta)}\cdot \sum_{t=1}^H\sum_{k=1}^K \Exs_{\theta^*}^{\pi^k} \left[ \|\bar{v}_{\theta^*}(\tau_t)\|_{\Lambda_*^k(x_t)^{-1}} \right] \\
    &\le \frac{M}{\alpha} \sqrt{(\lambda^* M + \beta) K} \cdot \sum_{t=1}^H \sqrt{\sum_{k=1}^K \Exs_{\theta^*}^{\pi^k} \left[ \|\bar{v}_{\theta^*}(\tau_t)\|_{\Lambda_*^k(x_t)^{-1}}^2 \right]}.
\end{align*}
Applying the expectation version of the elliptical potential lemma (see Lemma \ref{lemma:elliptical_potential_expectation}), by considering $\bar{v}_{\theta^*}(\tau_t)$ in the space of $\mathbb{R}^{MSA}$, and setting $\lambda^* = O(1)$, $\beta = \log(K|\Theta|/\delta) > M$, we have
\begin{align}
    \sum_{k=1}^K \TV(\PP_{\theta^*}^{\pi^k}(\tau, \iota), \PP_{\theta^k}^{\pi^k}(\tau, \iota)) &\lesssim \frac{MH}{\alpha} \sqrt{MSAK \beta \log(K)}, \label{main_eq:tv_distance_sum_bound}
\end{align}
with probability at least $1 - \delta$. Consequently, the regret bound is given by
\begin{align*}
    \sum_{k=1}^K V_{\theta^*}^{\pi_{\texttt{blind}}^*} - V_{\theta^*}^{\pi^k} &\lesssim \frac{M^{3/2}H^2}{\alpha} \sqrt{SAK \log(K|\Theta|/\delta) \log(K)},
\end{align*}
completing the proof.

\subsection{Proof of Lemma \ref{lemma:conditional_well-conditioned}}
\begin{proof}
    Recall that
    \begin{align*}
        \pi(\omega_t) \psi(\omega_t, \iota| x_t)^\top &= \pi(\omega_t) \cdot  \bm{e}_\iota^\top B(y_H|x_H) ... B(y_{t}|x_t) \\
        &= \mathbb{I}(\iota)^\top \diag(\PP^{\pi} (\omega_t | m, x_t)) \mathbb{I}^\dagger.
    \end{align*}
    Thus, 
    \begin{align*}
        \sum_{\omega_t} \pi(\omega_t) |\psi(\omega_t,\iota| x_t)^\top b| &= \sum_{\omega_t} |\mathbb{I}(\iota)^\top \diag(\PP^{\pi} (\omega_t | m, x_t)) \mathbb{I}^\dagger b| \\
        &\le \sum_{\omega_t} \sum_{m} |\PP(\iota|m) \PP^{\pi} (\omega_t | m,x_t)| \cdot |\bm{e}_m^\top \mathbb{I}^\dagger b | \\
        &\le \sum_{m} \PP(\iota|m) |\bm{e}_m^\top \mathbb{I}^\dagger b | \le \|\mathbb{I}(\iota)\|_{\infty} \|\mathbb{I}^\dagger b \|_1.
    \end{align*}
    Now applying Lemma G.4 in \cite{liu2023optimistic}, there exists a left-inverse of $\mathbb{I}$ such that $\|\mathbb{I}^\dagger b\|_1 \le M \|b\|_1 / \alpha$, and we have the result. 
\end{proof}

\subsection{Proof of Theorem \ref{theorem:reward_free_exploration}}
We divide the proof of this theorem into two parts. In the first part, we prove the required number of episodes until Algorithm \ref{algo:psr_ucb_reward_free} terminates. In the second part, we show the optimality of the returned model in a larger class of prospective side information exploiting policies $\Pi$. 

\subsubsection{Proof Part I}
The first part largely follows the proofs in \citet{huang2023provably} for the reward-free exploration until the sum of trajectory bonuses becomes small. The key step is connecting the trajectory bonuses between two different models in the confidence set. Define the bonus counterpart in the true environment:
\begin{align*}
    &\Lambda_t^k(x) = \lambda_0 I + \sum_{j<k} \indic{x_t^j = x} \bar{b}_{\theta^*}(\tau_t^j) \bar{b}_{\theta^*}(\tau_t^j)^\top, \\
    &\tilde{r}^k_*(\tau_t) = \|\bar{b}_{\theta^*}(\tau_t)\|_{\Lambda_t^k(x_t)^{-1}}.
\end{align*}
Then we compare that
\begin{align*}
     \|\bar{b}_{\theta^k}(\tau_t)\|_{\hat{\Lambda}_t^k(x_t)^{-1}} - \|\bar{b}_{\theta^*}(\tau_t)\|_{\Lambda_t^k(x_t)^{-1}}.
\end{align*}
Using Lemma \ref{lemma:bonus_comparison_lemma}, we can show that
\begin{align*}
    &\|\bar{b}_{\theta^k}(\tau_t)\|_{\hat{\Lambda}_t^k(x_t)^{-1}} - \|\bar{b}_{\theta^*}(\tau_t)\|_{\Lambda_t^k(x_t)^{-1}} \\
    &\le \frac{1}{\sqrt{\lambda_0}} \|\bar{b}_{\theta^k}(\tau_t) - \bar{b}_{\theta^*}(\tau_t)\|_2 \\
    &\quad + \|\bar{b}_{\theta^*}(\tau_t)\|_{\Lambda_t^k(x_t)^{-1}} \underbrace{\left \|\sum_{j < k} \indic{x_t^j = x_t} \hat{\Lambda}_t^k(x_t)^{-1/2} (\bar{b}_{\theta^*}(\tau_t^j) \bar{b}_{\theta^*}(\tau_t^j)^\top - \bar{b}_{\theta^k}(\tau_t^j) \bar{b}_{\theta^k}(\tau_t^j)^\top) \Lambda_t^k(x_t)^{-1/2} \right\|_2}_{(a)}.
\end{align*}
$(a)$ can be further bounded by
\begin{align*}
    (a) &\le \max_{u,v: \|u\|_2=1, \|v\|_2=1} \sum_{j < k} \indic{x_t^j = x_t} u \hat{\Lambda}_t^k(x_t)^{-1/2} (\bar{b}_{\theta^*}(\tau_t^j) \bar{b}_{\theta^*}(\tau_t^j)^\top - \bar{b}_{\theta^k}(\tau_t^j) \bar{b}_{\theta^k}(\tau_t^j)^\top) \Lambda_t^k(x_t)^{-1/2} v \\
    &\le \max_{u,v: \|u\|_2=1, \|v\|_2=1} \sum_{j < k} \indic{x_t^j = x_t} \left|u \hat{\Lambda}_t^k(x_t)^{-1/2} \bar{b}_{\theta^k}(\tau_t^j) \right| \left|(\bar{b}_{\theta^*}(\tau_t^j)^\top - \bar{b}_{\theta^k}(\tau_t^j))^\top \Lambda_t^k(x_t)^{-1/2} v \right| \\
    &\ + \max_{u,v: \|u\|_2=1, \|v\|_2=1} \sum_{j < k} \indic{x_t^j = x_t} \left|u \hat{\Lambda}_t^k(x_t)^{-1/2} (\bar{b}_{\theta^*}(\tau_t^j) - \bar{b}_{\theta^k}(\tau_t^j)) \right| \left|\bar{b}_{\theta^*}(\tau_t^j)^\top \Lambda_t^k(x_t)^{-1/2} v\right| \\
    &\le \sqrt{\sum_{j<k} \indic{x_t^j = x_t} \|\bar{b}_{\theta^k}(\tau_t^j)\|_{\hat{\Lambda}_t^k(x_t)^{-1} }^2 } \sqrt{\sum_{j<k} \indic{x_t^j = x_t} \|\bar{b}_{\theta^*}(\tau_t^j) - \bar{b}_{\theta^k}(\tau_t^j)\|_{\Lambda_t^k(x_t)^{-1}}^2 } \\
    &\ + \sqrt{\sum_{j<k} \indic{x_t^j = x_t} \|\bar{b}_{\theta^*}(\tau_t^j)\|_{\Lambda_t^k(x_t)^{-1} }^2 } \sqrt{\sum_{j<k} \indic{x_t^j = x_t} \|\bar{b}_{\theta^*}(\tau_t^j) - \bar{b}_{\theta^k}(\tau_t^j)\|_{\hat{\Lambda}_t^k(x_t)^{-1}}^2 } \\
    &\stackrel{(b)}{\le} \sqrt{\frac{M}{\lambda_0}} \sqrt{\sum_{j<k}  \|\bar{b}_{\theta^*}(\tau_t^j) - \bar{b}_{\theta^k}(\tau_t^j)\|_2^2 } \le \sqrt{\frac{M}{\lambda_0}} \sqrt{\sum_{j<k} \TV^2 (\PP_{\theta^*}^{\pi_j} (\iota | \tau_t^j), \PP_{\theta^k}^{\pi_j} (\iota | \tau_t^j))} \lesssim \sqrt{\frac{M\beta}{\lambda_0}},
\end{align*}
where for (b), we used Lemma \ref{lemma:trace_sum_bound}. 

Now taking expectation on both sides, we have
\begin{align*}
    \Exs_{\theta^*}^{\pi^k} \left[\|\bar{b}_{\theta^k}(\tau_t)\|_{\hat{\Lambda}_t^k(x_t)^{-1}} \right] &\le \left(1 + O(1) \cdot \sqrt{M\beta/\lambda_0}\right) \Exs_{\theta^*}^{\pi^k} \left[\|\bar{b}_{\theta^*}(\tau_t)\|_{\Lambda_t^k(x_t)^{-1}} \right] + \frac{O(1)}{\sqrt{\lambda_0}} \TV(\PP_{\theta^*}^{\pi_k}, \PP_{\theta^k}^{\pi_k}),
\end{align*}
where we used
\begin{align*}
    \Exs_{\theta^*}^{\pi^k} \left[\|\bar{b}_{\theta^k}(\tau_t) - \bar{b}_{\theta^*}(\tau_t)\|_2 \right] &\le \Exs_{\theta^*}^{\pi^k} \left[\|\bar{b}_{\theta^k}(\tau_t) - \bar{b}_{\theta^*}(\tau_t)\|_1 \right] \\
    &\le \Exs_{\theta^*}^{\pi^k} \left[\TV\left(\PP_{\theta^k}^{\pi^k} (\iota |\tau_t), \PP_{\theta^*}^{\pi^k}(\iota | \tau_t)\right) \right] \\
    &\le 2 \TV\left(\PP_{\theta^k}^{\pi^k} (\iota, \tau_t), \PP_{\theta^*}^{\pi^k}(\iota, \tau_t)\right).
\end{align*}
To proceed, note that $\|\bar{b}_{\theta^k}(\tau_t)\|_{\hat{\Lambda}_t^k(x_t) ^{-1}} \le \frac{1}{\sqrt{\lambda_0}}$ almost surely, and thus, 
\begin{align*}
    \Exs_{\theta^k}^{\pi^k} \left[\|\bar{b}_{\theta^k}(\tau_t)\|_{\hat{\Lambda}_t^k(x_t)^{-1}} \right] &\le \Exs_{\theta^*}^{\pi^k} \left[\|\bar{b}_{\theta^k}(\tau_t)\|_{\hat{\Lambda}_t^k(x_t)^{-1}} \right] + \frac{1}{\sqrt{\lambda_0}} \TV\left(\PP_{\theta^k}^{\pi^k}, \PP_{\theta^*}^{\pi^k}\right).
\end{align*}
Therefore, summing over $K$ episodes, we have
\begin{align*}
    \sum_{k=1}^K \Exs_{\theta^k}^{\pi^k} \left[\|\bar{b}_{\theta^k}(\tau_t)\|_{\hat{\Lambda}_t^k(x_t)^{-1}} \right] &\le \left(1 + O(1) \cdot \sqrt{M\beta/\lambda_0}\right) \sum_{k=1}^K \Exs_{\theta^*}^{\pi^k} \left[\|\bar{b}_{\theta^*}(\tau_t)\|_{\Lambda_t^k(x_t)^{-1}} \right] \\
    & \quad + \frac{O(1)}{\sqrt{\lambda_0}} \sum_{k=1}^K \TV\left(\PP_{\theta^k}^{\pi^k}, \PP_{\theta^*}^{\pi^k}\right).
\end{align*}
For the second term, we can apply equation \eqref{main_eq:tv_distance_sum_bound}. For the first term, we can first apply Azuma-Hoeffding inequality on
\begin{align*}
    \sum_{k=1}^K \left(\Exs_{\theta^*}^{\pi^k} \left[\|\bar{b}_{\theta^*}(\tau_t)\|_{\Lambda_t^k(x_t)^{-1}} \right] - \|\bar{b}_{\theta^*}(\tau_t^k)\|_{\Lambda_t^k(x_t)^{-1}} \right),
\end{align*}
and apply the empirical version of elliptical potential lemma (Lemma \ref{lemma:elliptical_potential_expectation}). This gives
\begin{align*}
    \sum_{k=1}^K \Exs_{\theta^k}^{\pi^k} \left[\|\bar{b}_{\theta^k}(\tau_t)\|_{\hat{\Lambda}_t^k(x_t)^{-1}} \right] &\lesssim \sqrt{MSAK \log(K)} \left(1 + O(1)\cdot \sqrt{M\beta/\lambda_0}+ (MH/\alpha) \cdot \sqrt{\beta/\lambda_0}\right).
\end{align*}
With the choice of $\lambda_0 = \frac{\beta M^2H^2}{\alpha^2}$, the Algorithm~\ref{algo:psr_ucb_reward_free} must terminate after at most $K$ episodes where
\begin{align*}
    K = O\left(\frac{MSA\log (K)}{\epsilon_{\texttt{pe}}^2}\right).
\end{align*}

\subsubsection{Proof Part II}
Now suppose Algorithm \ref{algo:psr_ucb_reward_free} terminated with the model $\theta$ that has the desired property:
\begin{align*}
    \max_{\pi\in\Pi_{\texttt{blind}}} V_{\theta, \tilde{r}}^{\pi} := \Exs_{\theta}^{\pi} \left[ \tssum_{t} \|\bar{b}_{\theta} (\tau_t) \|_{\hat{\Lambda}_t^k (x_t)^{-1}} \right] \le \epsilon_{\texttt{pe}}.
\end{align*}
Assuming this event holds true we continue the proof.
\begin{proof}    
We can express the total-variation distance between $\theta^*$ and $\theta$ as
\begin{align*}
    &\TV(\PP_{\theta^*}^\pi(\tau, \iota), \PP_{\theta}^\pi(\tau, \iota)) \le \sum_{t=1}^H \sum_{\tau} \pi(\tau) \cdot \left| \psi_{\theta^*}(\omega_{t+1}, \iota| x_{t+1})^\top \left(B_{\theta^*} (y_t|x_t) - B_{\theta} (y_t|x_t) \right) b_{\theta}(\tau_{t}) \right| \\
    &\le \sum_{\iota} \sum_{t=1}^H \sum_{\tau_t} \pi(\tau_t|\iota) \sum_{\omega_{t}} \pi(\omega_t | \tau_t, \iota) \cdot \left| \psi_{\theta^*}(\omega_{t+1}, \iota| x_{t+1})^\top \left(B_{\theta^*} (y_t|x_t) - B_{\theta} (y_t|x_t) \right) b_{\theta}(\tau_{t}) \right|.
\end{align*}
Notice that this time, we use $\theta^*$ to express the future prediction, and $\theta$ to express the history part in the above equation. Now we fix $\iota$, $t$ and $\tau_t$, and focus on bounding the inside summation. The first step is to normalize the belief state and rewrite the inner sum as:
\begin{align}
    &\sum_{\tau_t} \pi(\tau_t|\iota) \sum_{\omega_{t}} \pi^k(\omega_t | \iota, \tau_t) \cdot \left| \psi_{\theta^*}(\omega_{t+1}, \iota | x_{t+1})^\top \left(B_{\theta^*} (y_t|x_t) - B_{\theta} (y_t|x_t) \right) b_{\theta}(\tau_{t}) \right| \label{eq:intermediate_partII_1} \\
    &= \sum_{\tau_t} \pi(\tau_t|\iota) \|b_{\theta} (\tau_t)\|_1 \sum_{\omega_{t}} \pi^k(\omega_t | \iota, \tau_t) \cdot \left| \psi_{\theta^*}(\omega_{t+1}, \iota | x_{t+1})^\top \left(B_{\theta^*} (y_t|x_t) - B_{\theta} (y_t|x_t) \right) \bar{b}_{\theta}(\tau_{t}) \right|, \nonumber
\end{align}
where $\bar{b}_{\theta} (\tau_{t}) = \frac{b_{\theta}(\tau_t)}{\| b_{\theta}(\tau_t)\|_1} $ are the normalized predictive representation of belief states. Then note that $\pi(\tau_t|\iota) \| b_{\theta}(\tau_t) \|_1 = \PP_{\theta^*}^{\pi(\cdot|\iota)} (\tau_t)$, \textbf{\emph i.e., a marginalized probability of $\tau_t$ when running a prospective side information blind policy} $\pi(\cdot|\iota)$: 
\begin{align*}
    \PP_{\theta^*}^{\pi(\cdot|\iota)} (\tau_t) &= \sum_{\iota'} \PP_{\theta^*}^{\pi(\cdot|\iota)} (\tau_t, \iota'),
\end{align*}
%\yecomment{unfold the experssion here if not too long (o.w., add a short lemma?)} 
as if we do not use the true prospective side information but instead use an arbitrary dummy variable $\iota$ to instantiate a blind policy. Thus, we can express \eqref{eq:intermediate_partII_1} as
\begin{align*}
    \Exs_{\tau_t \sim \PP_{\theta}^{\pi(\cdot|\iota) }(\cdot)} \left[ \sum_{\omega_t} \pi(\omega_t | \iota, \tau_t) \left| \psi_{\theta^*} (\omega_{t+1}, \iota | x_{t+1})^\top \left(B_{\theta^*} (y_t|x_t) - B_{\theta} (y_t|x_t) \right) \bar{b}_{\theta} (\tau_{t}) \right| \right].
\end{align*}
Recall the empirical pseudo-count matrix:
\begin{align*}
    \hat{\Lambda} (s,a) &= \lambda_0 I + \sum_{k \in [K]} \left[\indic{(s_t^k, a_t^k) = (s,a)} \cdot \bar{b}_{\theta}(\tau_t^k) \bar{b}_{\theta}(\tau_t^k)^\top \right].
\end{align*} 
For simplicity, let $f(\omega_t) := \psi_{\theta^*}(\omega_{t+1}, \iota | x_{t+1})^\top \left(B_{\theta^*} (y_t| x_t) - B_{\theta} (y_t| x_t) \right)$ ($f$ is only a function of $\omega_t$ as other variables are fixed at this point). Using Cauchy-Schwartz inequality, we have
\begin{align*}
    \eqref{eq:intermediate_partII_1} &\le \Exs_{\tau_t \sim \PP_{\theta}^{\pi(\cdot|\iota)}} \left[ \left\| \sum_{\omega_t} \pi(\omega_t | \iota, \tau_t) f(\omega_t) \cdot \sign(f(\omega_t)^\top \bar{b}_{\theta}(\tau_t)) \right\|_{\hat{\Lambda} (x_t)} \|\bar{b}_{\theta} (\tau_{t})\|_{\hat{\Lambda} (x_t)^{-1}} \right].
\end{align*}
To bound the concentration bound, we can check that
\begin{align}
    &\left\| \sum_{\omega_t} \pi(\omega_t | \iota, \tau_t) f(\omega_t) \cdot \sign(f(\omega_t)^\top \bar{b}_{\theta}(\tau_t)) \right\|_{\hat{\Lambda} (x_t)}^2 \nonumber \\
    &\le \lambda_0 \left\| \sum_{\omega_t} \pi(\omega_t | \iota, \tau_t) f(\omega_t) \cdot \sign(f(\omega_t)^\top \bar{b}_{\theta}(\tau_t)) \right\|_2^2 \nonumber \\
    &+ \sum_{k \in [K]} \indic{x_t^k = x_t} \left( \sum_{\omega_t} \pi(\omega_t | \iota, \tau_t) \left(f(\omega_t)^\top \bar{b}_{\theta}(\tau_t^k) \right) \cdot \sign(f(\omega_t)^\top \bar{b}_{\theta}(\tau_t)) \right)^2. \label{eq:PartII_concentration_bound}
\end{align}
For the term with $\lambda_0$, note that any vector $v$ that lies on the orthogonal complement of the span of $\mathbb{I}_{\theta}$, $\mathbb{I}_{\theta}^{\dagger} v = 0$. %\yecomment{introduce equation number here (and in other places in which there is no direct continuation)} 
Consider a vector $v = \mathbb{I}_{\theta} u$ such that $\|\mathbb{I}_{\theta} u\|_2 \le 1$. Note that to satisfy this condition, $u$ cannot be too large: $\|u\|_1 \le \max_{\|v\|_2 = 1} \|\mathbb{I}_{\theta}^{\dagger} v\|_1 \le \max_{\|v\|_1 = 1} \|\mathbb{I}_{\theta}^{\dagger} v\|_1 \le \frac{M}{\alpha}$. Thus, 
\begin{align*}
    &\left| \sum_{\omega_t} \pi(\omega_t | \iota, \tau_t) f(w_t) v \right| \\
    &\le \sum_{\omega_t} \pi(\omega_t | \iota, \tau_t) | \psi_{\theta^*}(\omega_{t+1}, \iota | x_{t+1})^\top \left(B_{\theta^*} (y_t| x_t) - B_{\theta} (y_t| x_t) \right) \mathbb{I}_{\theta} u) | \\
    &\le \sum_{\omega_t} \pi(\omega_t | \iota, \tau_t) | \psi_{\theta^*}(\omega_{t}, \iota | x_{t})^\top \mathbb{I}_{\theta} u - \psi_{\theta^*}(\omega_{t+1}, \iota | x_{t+1})^\top \mathbb{I}_{\theta} \diag(\PP_{\theta}(y_t | m,x_t)) u  | \\
    &\le \frac{2M}{\alpha} \|\mathbb{I}_{\theta^*}(\iota)\|_{\infty} \|u\|_1 = \frac{2M^2}{\alpha^2} \|\mathbb{I}_{\theta^*}(\iota)\|_{\infty},
\end{align*}
where we applied Lemma \ref{lemma:mle_traj_concentration}, and therefore 
\begin{align*}
    \left\| \sum_{\omega_t} \pi(\omega_t | \iota, \tau_t) f(w_t) \right\|_2^2 \le \frac{4M^4}{\alpha^4} \|\mathbb{I}_{\theta^*}(\iota)\|_{\infty}^2.
\end{align*}

To bound the second term in \eqref{eq:PartII_concentration_bound}, first we  observe the term inside the summation (over $k$) is only nonzero when $x_t^k=x_t$, i.e., $(s_t^k,a_t^k) = (s_t,a_t)$. We have that
\begin{align*}
    &\sum_{\omega_t} \pi(\omega_t | \iota, \tau_t) \left(f(\omega_t)^\top \bar{b}_{\theta}(\tau_t^k) \right) \cdot \sign(f(\omega_t)^\top \bar{b}_{\theta}(\tau_t)) \\
    &\le \sum_{\omega_t} \pi(\omega_t | \iota, \tau_t) \left| f(\omega_t)^\top \bar{b}_{\theta}(\tau_t^k)  \right| \\
    &= \sum_{\omega_t} \pi(\omega_t | \iota, \tau_t) \left| \psi_{\theta^*}(\omega_{t+1}, \iota | x_{t+1})^\top \left( B_{\theta^*} (y_t | x_t) - B_{\theta} (y_t| x_t) \right) \bar{b}_{\theta}(\tau_t^k)  \right| \\
    &\le \sum_{\omega_t} \pi(\omega_t | \iota, \tau_t) \left| \psi_{\theta^*}(\omega_{t+1},\iota | x_{t+1})^\top \left( B_{\theta^*} (y_t | x_t)\bar{b}_{\theta^*}(\tau_t^k) - B_{\theta} (y_t| x_t) \bar{b}_{\theta}(\tau_t^k) \right)  \right| \\
    &\quad + \sum_{\omega_t} \pi(\omega_t | \iota, \tau_t) \left| \psi_{\theta^*}(\omega_{t},\iota | x_t)^\top \left( \bar{b}_{\theta^*}(\tau_t^k) - \bar{b}_{\theta}(\tau_t^k) \right) \right| \\
    &\le \frac{M}{\alpha} \|\mathbb{I}_{\theta^*} (\iota)\|_{\infty} \cdot \Bigg( \sum_{y_t} \| B_{\theta} (y_t | x_t)\bar{b}_{\theta}(\tau_t^k) - B_{\theta^*} (y_t|x_t) \bar{b}_{\theta^*}(\tau_t^k) \|_1 + \| \bar{b}_{\theta}(\tau_t^k) - \bar{b}_{\theta^*}(\tau_t^k) \|_1 \Bigg),
\end{align*}
where we denote $\bar{b}_{\theta} = \mathbb{I}_{\theta} \bar{b}_{\theta}$. 
We can check the meaning of each term: for any $\iota' \in\mI$ and any {\it blind} policy $\pi \in \Pi_{\texttt{blind}}$, 
\begin{align*}
    &\indic{x_t^k = x_t} \cdot \bm{e}_{\iota'}^\top B_{\theta} (y_t |x_t) \bar{b}_{\theta}(\tau_t^k) \\
    &= \frac{\mathds{1}^\top \diag(\PP_{\theta} ( \iota' | m)) \Pi_{h=1}^{t} \diag(\PP_{\theta} (y_{h} |m,x_{h}^k )) w }{ \| v_{\theta}(\tau_t^k) \|_1 } \\
    &= \frac{\mathds{1}^\top \diag(\PP_{\theta} ( \iota' | m)) \Pi_{h=1}^{t} \diag(\PP_{\theta} (y_{h} |m,x_{h}^k )) w}{\sum_{\iota''} \mathds{1}^\top \diag(\PP_{\theta} (\iota''|m)) \Pi_{h=1}^{t-1} \diag_{\theta} (\PP(y_{h} |m,x_{h}^k)) w} \\
    &=  \frac{\pi(\tau_t^k) \cdot \mathds{1}^\top \diag(\PP( \iota' | m)) \Pi_{h=1}^{t} \diag(\PP(y_{h} |m,x_{h}^k)) w}{\sum_{\iota''} \pi(\tau_t^k) \cdot \mathds{1}^\top \diag(\PP(\iota''|m)) \Pi_{h=1}^{t-1} \diag(\PP(y_{h} |m,x_{h}^k)) w} \\
    &= \frac{\PP^{\pi}_{\theta} (\iota', y_{t}, \tau_t^k)}{\PP^{\pi}_{\theta} (\tau_t^k)} = \PP^{\pi}_{\theta} (i', y_t | \tau_t^k).
\end{align*}
To proceed, let the prospective side information blind policy executed on the $k^{th}$ episode be $\pi^{k}$. We have that
\begin{align*}
    &\sum_{\omega_t} \indic{x_t^k = x_t} \pi(\omega_t | \iota, \tau_t) \left(f(\omega_t)^\top \bar{b}_{\theta}(\tau_t^k) \right) \cdot \sign(f(\omega_t)^\top \bar{b}_{\theta}(\tau_t)) \\
    &\le \frac{2 M}{\alpha} \|\mathbb{I}_{\theta^*} (\iota)\|_{\infty} \indic{x_t^k = x_t} \cdot \TV \left(\PP^{\pi^{k}}_{\theta^*} (\iota', y_t | \tau_t^k), \PP^{\pi^{k}}_{\theta} (\iota', y_t | \tau_t^k) \right).
\end{align*}

Combining the result, we conclude that
\begin{align*}
    \eqref{eq:intermediate_partII_1} &\le \Exs_{\tau_t \sim \PP_{\theta}^{\pi(\cdot|\iota) } } \left[ \left\| \sum_{\omega_t} \pi(\omega_t | \iota, \tau_t) f(\omega_t) \cdot \sign(f(\omega_t)^\top \bar{b}_{\theta}(\tau_t)) \right\|_{\hat{\Lambda}(x_t)} \|\bar{b}_{\theta} (\tau_{t})\|_{\hat{\Lambda} (x_t)^{-1}} \right] \\
    &\le \Exs_{\tau_t \sim \PP_{\theta}^{\pi(\cdot|\iota) }(\cdot)} \left[ \frac{2M}{\alpha} \|\mathbb{I}_{\theta^*} (\iota)\|_{\infty} \cdot c(x_t) \|\bar{b}_{\theta} (\tau_{t})\|_{\hat{\Lambda} (x_t)^{-1}} \right],
\end{align*}
where
\begin{align*}
    c(x_t) &= \sqrt{\frac{M^2}{\alpha^2} \lambda_0 + \sum_{k \in [K]} \left[ \indic{x_t^k = x_t} \TV^2 \left(\PP^{\pi^{k} }_{\theta^*} (\iota', y_t' | \tau_t^k), \PP^{ \pi^{k} }_{\theta} (\iota', y_t' | \tau_t^k) \right) \right] } \\
    &\le \sqrt{\frac{M^2}{\alpha^2} \lambda_0 + \sum_{k \in [K]} \TV^2 \left(\PP^{\pi^{k} }_{\theta^*} (\iota', y_t' | \tau_t^k), \PP^{ \pi^{k} }_{\theta} (\iota', y_t' | \tau_t^k) \right) } \\
    &\lesssim \sqrt{\frac{M^2}{\alpha^2} \lambda_0 + \beta} := c_{\max},
\end{align*}
where in the second inequality, we used Lemma \ref{lemma:concentration_conditioanl_tv_dist}. Now proceeding,
\begin{align*}
     &\sum_{\iota, \tau_H} \left|\PP_{\theta^*}^{\pi} (\iota, \tau_H) - \PP_{\theta}^{\pi} (\iota, \tau_H) \right| \\
     &\le \frac{2M}{\alpha} c_{\max}  \sum_{\iota}  \|\mathbb{I}_{\theta^*} (\iota)\|_{\infty} \sum_t \Exs_{\tau_t \sim \PP_{\theta}^{\pi(\cdot|\iota)}} \left[ \|\bar{b}_{\theta} (\tau_{t})\|_{\hat{\Lambda} (x_t)^{-1}} \right] \\
     &= \frac{2M}{\alpha} c_{\max} \sum_{\iota}  \|\mathbb{I}_{\theta^*} (\iota)\|_{\infty} \cdot \Exs_{\theta}^{\pi(\cdot|\iota)} \left[ \sum_t \|\bar{b}_{\theta} (\tau_{t})\|_{\hat{\Lambda} (x_t)^{-1}} \right] \\
     &\le \frac{2M}{\alpha} c_{\max} \sum_{\iota}  \|\mathbb{I}_{\theta^*} (\iota)\|_{\infty} \cdot \max_{\pi \in \Pi_{\text{blind}}} \Exs_{\theta}^\pi \left[ \sum_t \|\bar{b}_{\theta} (\tau_{t})\|_{\hat{\Lambda} (x_t)^{-1}} \right] \\
     &\stackrel{(a)}{\le} \frac{2M^2}{\alpha} \sqrt{\frac{M^2 \lambda_0}{\alpha^2} + \beta} \cdot \epsilon_{\texttt{pe}},
\end{align*}
where $(a)$ comes from $\sum_{\iota}  \|\mathbb{I}_{\theta^*} (\iota)\|_{\infty} \le \sum_{m} \sum_{\iota} \mathbb{I}_{\theta^*} (m, \iota) = M$. With the choice of $\lambda_0 = \beta M^2 H^2 / \alpha^2$, by setting $\epsilon_{\texttt{pe}} := \frac{\alpha \epsilon}{10 H M^2 \sqrt{M^4 H^2 \beta / \alpha^4}}$ ensures that 
\begin{align*}
    | V^{\pi}_{\theta} - V^{\pi}_{\theta^*} | \le H \TV(\PP_{\theta}^{\pi}, \PP_{\theta^*}^{\pi}) \le \epsilon/2,
\end{align*}
for all $\pi \in \Pi$. Therefore, optimizing over $\theta$ gives $\epsilon$-optimal policy for $\theta^*$, completing the proof. 
\end{proof}

\section{Lower Bound Proofs}
\label{appendix:proof:lower_bound}
We first complete the construction of the hard instance family deferred from the main text. Recall that we defined:
\begin{enumerate}
    \item Action space: 
        \begin{itemize}
            \item $\mA_{\text{exploit}} = \{a_{\text{exploit}}^m\}$ for $m=M/2+1, ..., M$
            \item $\mA_{\text{explore}}$: contains the true exploration action (at the initial step) $a_{\text{explore}}^*$ and dummy actions
            \item $\mA_{\text{control}}$: contains the optimal actions $a_{t}^*$ for $t \in [d]$ and dummy actions
        \end{itemize}
    \item State space: 
        \begin{itemize}
            \item $s_{\text{init}}, s_{\text{ter}}$: initial and terminated state
            \item $s_{1:d}^{\text{hard}}$: chained states in the hard-to-learn chain
            \item $s_{1:d}^{\text{ref}}$: chained states in the reference chain
        \end{itemize}
    \item MDP groups in an LMDP: 
        \begin{itemize}
            \item $\mG_{\text{learn}}$: a set of MDPs that needs to be acted optimally in the hard-to-learn chain
            \item $\mG_{\text{ref}}$: a set of MDPs that confuses the identity of hard-to-learn and reference chains
            \item $\mG_{\text{obs}}$: a set of MDPs whose identity is strongly correlated to the prospective side information
        \end{itemize}
\end{enumerate}

\paragraph{Initial Transition Setup.} $\mG_{\texttt{learn}}$ consists of $(M/4)$ MDPs , $\mM_1, ..., \mM_{M/4}$, which form the hard-to-learn example from \citet{kwon2021rl} when no prospective side information is provided. In any of these MDPs in $\mG_{\texttt{learn}}$, at the beginning, if an action $a_{\text{explore}}^*$ is executed, the environment transitions to the starting of the hard-instance chain $s^{\text{hard}}_{1}$ with probability $\epsilon > 0$, or transitions to the starting of reference chain $s^{\text{ref}}_{1}$ with probability $1 - \epsilon$. If any other action in $\mA_{\text{explore}}$ is executed, the environment transitions 
 to $s^{\text{ref}}_1$ with probability 1. For all other actions executed, the MDPs  transition to a terminate-state $s_{\text{ter}}$.

$\mG_{\texttt{ref}}$ consists of another $(M/4)$ MDPs, $\mM_{M/4+1}, ..., \mM_{M/2}$, which suppose to confuse the learning process in $\mG_{\texttt{learn}}$. In these environments, dynamics in hard-to-learn chain and the reference chain are the same. Instead, at the beginning of an episode, if $a_{\text{explore}}^*$ is executed, an MDP transitions to the starting of hard-to-learn chain $s^{\text{hard}}_{1}$ with probability $1 - \epsilon$, or transitions to the starting of reference chain $s^{\text{ref}}_1$ with probability $\epsilon$. If any other action in $\mA_{\text{explore}}$ is executed, the environment transitions to $s^{\text{hard}}_1$ with probability 1. Like the group $\mG_{\texttt{learn}}$, for all other actions executed, the MDPs in $\mG_{\texttt{ref}}$ transition to $s_{\text{ter}}$. 

The of the MDPs, indexed by $\mM_{M/2+1}, ..., \mM_M$, belong to the almost observable group $\mG_{\texttt{obs}}$. In any of these MDPs in $\mG_{\texttt{obs}}$, the environment always transitions to an absorbing state $s_{\text{ter}}$ after playing an initial action. In each environment of this group $\mM_m \in \mG_{\texttt{obs}}$ where $m=M/2+1,...,M$, playing $a^{m}_{\text{exploit}}$ results with a reward of 1, and with 0 when playing any action different than $a^{m}_{\text{exploit}}$.

\paragraph{Prospective Side Information Setup.} The prospective side information is a finite alphabet  belongs and belongs to one of the $M+1$ disjoint sets $\mI_1, \mI_2, ..., \mI_M, \mI_{M+1}$. We let $\mI_{M+1} := \{ \iota_{\text{hard}} \}$ contains a single element, and all other disjoint sets have equal cardinality $|\mI_1| = |\mI_2| = ... = |\mI_M|:= |\mI|.$ % Note that the size of $\mI = \cup_{i=1}^{M+1} \mI_i$ can be arbitrarily large. 
For $\mM_{M/2+1}, ..., \mM_{M}$ in $\mG_{\texttt{obs}}$, for each $m \in [M/2+1, M]$, the emission probability is give by $\PP(\iota|m) = 1/(2|\mI|)$ if $\iota \in \mI_{m-M/2} \cup \mI_{m}$, and $0$ otherwise. 

For all environments $\mM_{1}, ..., \mM_{M/4} \in \mG_{\texttt{learn}}$ and $\mM_{M/4+1}, ..., \mM_{M/2} \in \mG_{\texttt{ref}}$, for all $m \in \left[M/2\right]$,  $\PP(\iota_{\text{hard}} | m) = 1/2$, and $\PP(\iota | m) = 0$ if $\iota \in \bigcup_{i=M/2+1}^{M} \mI_i$. For all $\iota \in \bigcup_{i=1}^{M/2} \mI_i$, we assign the probability of prospective side information $\PP(\iota|m) \propto \frac{1 + \alpha \varepsilon_{\iota,m}} {M |\mI_m|}$, where each $\varepsilon_{\iota,m} \in \{-1,1\}$ is decided in the following lemma: 
\begin{lemma}
    \label{lemma:probabilistic_separation}
    There exists a set of $\{\varepsilon_{\iota,m}\}_{\iota, m}$ such that for all $x \in \mathbb{R}^M$ it holds that $\|x\|_1 = 1$, $\| \mathbb{I} x \|_1 \ge \alpha' = \frac{\alpha}{128 \sqrt{M}}$. 
\end{lemma}

\paragraph{Construction of Hard-to-Learn Chain for $\mG_{\texttt{learn}}$, where $\mM_{1}, ..., \mM_{d}$ with $d = M/4$.} This set is also depicted in the top part of Figure~\ref{fig:hard_instance}. 
\begin{itemize}
    \item At $t = 1$, {\it i.e.,} $s_1^{\text{hard}}$, there are three state-transition possibilities:
    \begin{itemize}
        \item $\mM_1$: For all actions $a \in \mA_{\text{control}}$ except $a_1^*$, we go to $s_{\text{ter}}$. For the action $a_1^*$, we go to $s_{2}^{\text{hard}}$.
        \item $\mM_d$: For all actions $a \in \mA_{\text{control}}$ except $a_1^*$, we go to $s_{2}^{\text{hard}}$. For the action $a_1^*$, we go to $s_{\text{ter}}$.
        \item $\mM_2$, ..., $\mM_{d-1}$: For all actions $a \in \mA_{\text{control}}$, we go to $s_{2}^{\text{hard}}$.
    \end{itemize}
    
    \item At time step $t = 2$, we again have three cases but now $\mM_1$ and $\mM_{d}$ would look the same:
    \begin{itemize}
        \item $\mM_1, \mM_d$: For all actions $a \in \mA_{\text{control}}$ except $a_2^*$, we go to $s_{\text{ter}}$. For the action $a_2^*$, we go to $s_{3}^{\text{hard}}$.
        \item $\mM_{d-1}$: For all actions $a \in \mA_{\text{control}}$ except $a_2^*$, we go to $s_{3}^{\text{hard}}$. For the action $a_2^*$, we go to $s_{\text{ter}}$.
        \item $\mM_2$, ..., $\mM_{d-2}$: For all actions $a \in \mA_{\text{control}}$, we go to $s_{3}^{\text{hard}}$.
    \end{itemize}
    ...
    \item[d.] At time step $t = d$, we always transitionto $s_{\text{ter}}$, and there are two possibilities of getting rewards:
    \begin{itemize}
        \item $\mM_1$: For the action $a_{d}^* \in \mA_{\text{control}}$, we get reward 1. For all other actions, we get rewards from $\Ber(1/8)$.
        \item $\mM_2, ..., \mM_d$: For all actions $a \in \mA_{\text{control}}$, we get rewards from $\Ber(1/8)$.
    \end{itemize}
\end{itemize}

\subsection{Proof of Lemma \ref{lemma:probabilistic_separation}}
\label{appendix:proof:probabilistic_separation}
\begin{proof}
    This can be shown by probabilistic arguments. Note that prospective side information that belongs to $\bigcup_{i=1}^{M/2} \mI_i$ uniquely identifies the environment from $\mG_{\texttt{no}}$, and thus
    \begin{align*}
        \|\mathbb{I} x\|_1 &= \|\mathbb{I}_{1:M/2} x\|_1 + \|\mathbb{I}_{M/2+1:M+1} x\|_1 \\
        &\ge \frac{1}{2} \|x_{M/2+1:M}\|_1 + \max\left(0, \|\mathbb{I}_{M/2+1:M+1} x_{1:M/2}\|_1 - \frac{1}{2} \|x_{M/2+1:M}\|_1\right),
    \end{align*}
    where with slight abuse in notation, we denote $\mathbb{I}_{i:j}$ as the sub-matrix whose rows only correspond to one of prospective side information groups $\mI_i, \mI_{i+1}, ..., \mI_{j}$. It is easy to check that if $\|\mathbb{I}_{M/2+1:M+1} x_{1:M/2}\|_1 \ge \frac{\alpha}{64\sqrt{M}} \|x_{1:M/2}\|_1$, then \begin{align*}
        \|\mathbb{I} x\|_1 &\ge \frac{1}{2} \|x_{M/2+1:M}\|_1 + \max\left(0, \frac{\alpha}{\sqrt{M}} \|x_{1:M/2}\|_1 - \frac{1}{2}\|x_{M/2+1:M}\|_1\right) \\
        &\ge \frac{1}{2} \|x_{M/2+1:M}\|_1 + \max\left(0, \frac{\alpha}{\sqrt{M}} - \|x_{M/2+1:M}\|_1\right) \\
        &\ge \frac{\alpha}{128 \sqrt{M}}. 
    \end{align*} 
    Thus, it is sufficient to show that there exists $\{\varepsilon_{\iota, m}\}_{\iota,m}$ such that
    \begin{align*}
        \|\mathbb{I}_{M/2+1:M+1} x_{1:M/2}\|_1 \ge \|\mathbb{I}_{M/2+1:M} x_{1:M/2}\| \ge \frac{\alpha}{64 \sqrt{M}} \|x_{1:M/2}\|_1.
    \end{align*}
    
    \paragraph{Probabilistic Assignment.} We set each $\varepsilon_{\iota,m}$ by an independent uniform sampling over $\{-1,1\}$. We assume that $|\mI_m|$ is sufficiently large, so that $\sum_{\iota \in \mI} \varepsilon_{\iota,m}$ concentrates around $0$ within $1/\sqrt{|\mI|}$ and $1/\sqrt{|\mI|}$ is sufficiently small.

    \paragraph{Probabilistic Existence.} To simplify the notation, we let $\mathbb{J} = \|\mathbb{I}_{M/2+1:M} x_{1:M/2}\|$ and $v = x_{1:M/2}$. Consider an $\gamma = \frac{\alpha}{256\sqrt{M}}$-cover, $\mathbb{B}_{\gamma}$  for the set $\{v \in \mathbb{R}^{M/2}: \|v\|_1 = 1\}$. Note that for each row of $\mathbb{J}$ and each $v \in \mathbb{B}_{\gamma}$, 
    \begin{align*}
        |\mathbb{J}_\iota^\top v| = \frac{1}{M |\mI|} \left| \tssum_{m\in[M/2]} v_m + \alpha \cdot \sum_{m\in[M/2]} v_m \varepsilon_{\iota,m} \right|. 
    \end{align*}
    Without loss of generality, we assume $\tssum_{m\in[M/2]} v_m \ge 0$. Note that the statistics of $W:= |\sum_{m\in[M/2]} v_m \varepsilon_{\iota,m}|$, by Paley–Zygmund inequality \cite{paley1930some},
    \begin{align*}
        \PP \left(W \ge \frac{1}{2} \|v\|_2 \right) \ge \frac{3}{16}, 
    \end{align*}
    and thus with probability at least $3/32$, we have
    \begin{align*}
        \sum_{m\in[M/2]} v_m \varepsilon_{\iota,m} \ge \frac{1}{2\sqrt{M}} \implies |\mathbb{J}_{\iota}^\top v| \ge \frac{\alpha}{2\sqrt{M}}.
    \end{align*}
    Since this holds for each row, and all $\epsilon_{\iota,m}$ are independent across the rows, at least $\frac{3}{64} (M/2) |\mI_1|$ rows satisfies the above with probability at least $1 - \exp(-(M/8) |\mI_1|)$ from the concentration of the sum of independent Bernoulli random variables, which translates to 
    \begin{align*}
        \|\mathbb{J}^\top v\|_1 \ge \frac{3 \alpha}{128 \sqrt{M}}, 
    \end{align*}
    with probability $1 - \exp(-(M/8) |\mI_1|)$. Therefore, taking a union bound over $\mathbb{B}_{\gamma}$, we have
    \begin{align*}
        \|\mathbb{J}^\top v\|_1 \ge \frac{3 \alpha}{128 \sqrt{M}},
    \end{align*}
    with probability $1 - |\mathbb{B}_{\gamma}| \exp(-(M/8)|\mI_1|) \ge 1 - \exp(c_1 M \log(\gamma) - c_2 |\mI|)$ with proper absolute constants $c_1, c_2 > 0$. Then for arbitrary $v: \|v\|_1 = 1$, we can always find $v_{\gamma}$ in $\mathbb{B}_{\gamma}$ such that $\|v - v_{\gamma}\| \le \gamma$, and therefore
    \begin{align*}
        \|\mathbb{J}^\top v\|_1 \ge \|\mathbb{J}^\top v_{\gamma}\|_1 - \|\mathbb{J}^\top (v - v_{\gamma}) \|_1 \ge \frac{3\alpha}{128 \sqrt{M}} - M \gamma.
    \end{align*}
    Thus, setting $\gamma = o(\alpha / \sqrt{M})$ sufficiently small, for all $v: \|v\|_1 = 1$, 
    \begin{align*}
        \|\mathbb{J}^\top v\|_1 \ge \frac{\alpha}{64\sqrt{M}}. 
    \end{align*}
    Since this probabilistic argument implies the existence of $\{\epsilon_{\iota,m}\}$, the proof is done. 
\end{proof}

\subsection{Proof of Lemma \ref{lemma:information_equality}}
This comes from the fundamental equality for sequential decision making information gain (see {\it e.g.,} \citet{cesa2006prediction, garivier2019explore, kwon2023reward}). For completeness, we prove this. We can start from
\begin{align*}
    \KL &\left(\PP_{\theta_0}^\psi (\tau^{1:K}), \PP_{\theta}^{\psi} (\tau^{1:K}) \right) = \Exs_{\theta_0} \left[\log \left( \frac{\PP_{\theta_0}^{\psi} (\tau^{1:K-1})}{\PP_{\theta}^{\psi} (\tau^{1:K-1})} \right) \right] + \Exs_{\theta_0} \left[\log \left( \frac{\PP_{\theta_0}^{\psi} (\tau^K | \tau^{1:K-1})}{ \PP_{\theta}^{\psi} (\tau^K | \tau^{1:K-1})} \right) \right].
\end{align*}
Note that in all models in our construction set $\Theta_{\text{hard}} \cup \{\theta_0\}$, $\PP(\iota)$ and $\psi(a_t^k| \text{all histories until $k^{th}$ episode, $t^{th}$ step})$ are the same. Therefore, we have that
\begin{align*}
    \Exs_{\theta_0} &\left[\log \left( \frac{\PP_{\theta_0}^{\psi} (\tau^K | \tau^{1:K-1})}{ \PP_{\theta}^{\psi} (\tau^K | \tau^{1:K-1})} \right) \right] \\
    &= \Exs_{\theta_0}^{\psi} \left[ \Exs_{\theta_0}^{\psi} \left[ \sum_{\iota, a, a_{1:d}} \log \left( \frac{\PP_{\theta_0}^{\psi} \left(\cdot | \iota, a, a_{1:d} \right)}{\PP_{\theta}^{\psi} \left(\cdot | \iota, a, a_{1:d} \right)} \right) \indic{(\iota, a, a_{1:d})^K = (\iota, a, a_{1:d})} \Big| \tau^{1:K-1} \right] \right] \\
    &= \sum_{\iota, a, a_{1:d}} \Exs_{\theta_0}^{\psi} \left[ \Exs_{\theta_0}^{\psi} \left[ \log \left( \frac{\PP_{\theta_0}^{\psi} \left(\cdot | \iota, a, a_{1:d} \right)}{\PP_{\theta}^{\psi} \left(\cdot | \iota, a, a_{1:d} \right)} \right) \Big| \iota, a, a_{1:d} \right] \indic{(\iota, a, a_{1:d})^K = (\iota, a, a_{1:d})}  \right] \\
    &= \sum_{\iota, a, a_{1:d}} \KL\left( \PP_{\theta_0} (\cdot | \iota, a, a_{1:d}), \PP_{\theta} (\cdot | \iota, a, a_{1:d}) \right) \cdot \Exs_{\theta_0}^{\psi} \left[\indic{(\iota, a, a_{1:d})^K = (\iota, a, a_{1:d})}  \right],
\end{align*}
where the second equality is an application of the tower rule, and the last equality is due to the choice of action purely depends on the history and exploration strategy $\psi$, and does not depend on underlying models. Applying this recursively in $K$, and denoting $N_{\psi, \iota, a_{1:d}}^a (K)$ as the number of times  action $a$ was executed at the initial step, $a_{1:d}$ in the next $d$ steps under prospective side information $\iota$. Thus, we have
\begin{align*}
    \KL \left(\PP^{\psi}_{\theta_0} (\tau^{1:K}), \PP^{\psi}_{\theta} (\tau^{1:K}) \right) &= \sum_{\iota, a, a_{1:d}} \Exs_{\theta_0} \left[ N_{\psi, \iota, a_{1:d}}^a (K) \right] \cdot \KL\left( \PP_{\theta_0} (\cdot | \iota, a, a_{1:d}), \PP_{\theta} (\cdot | \iota, a, a_{1:d}) \right),
\end{align*}
Note that when playing $a \neq a_{\text{explore}}^*$, the hard instance and the reference model behave the same, yielding the result.

\subsection{Proof of Lemma \ref{lemma:kl_divergence_bound}}
We first check the following inequality:
\begin{align*}
    \KL\left( \PP_{\theta_0} (\cdot | \iota_{\text{hard}}, a_{\text{explore}}^*, a_{1:d}^* ), \PP_{\theta} (\cdot | \iota_{\text{hard}}, a_{\text{explore}}^*, a_{1:d}^* ) \right).
\end{align*}
The point is that until seeing the last time-step event, the distribution of histories are the same in all environments. To see this, at the initial time step given the prospective side information $\iota_{\text{hard}}$, the belief over latent contexts are all equal to $2/M$ for all MDPs in $\mG_{\texttt{learn}}$ and $\mG_{\texttt{ref}}$. Thus, the probability of transitioning to $s_1^{\text{hard}}$ is $1/2$ by executing $a^*_{\text{explore}}$ (if the environment transitions to $s_1^{\text{ref}}$, or any other action is executed, then the future distribution on of all events are exactly the same in all hard and reference instances). In the middle of the hard-instance chain, at $s_{t}^{\text{hard}}$, the probability of moving to the next state conditioned on the past is $1 - 1/(d - t + 1)$. However, the true posterior probability over MDPs from $\mG_{\texttt{ref}}$ at this point is given by:
\begin{align*}
    \PP(m | \iota_{\text{hard}}, a_{1:t}, s_{t}^{\text{hard}}) = \epsilon / (d - t + 1),
\end{align*}
for all $m = 1, 2, ..., M/4$ with non-zero posteriors (since we eliminated MDPs from the set after gathering information in a certain way). On the other hand, 
\begin{align*}
    \PP(m | \iota_{\text{hard}}, a_{1:t}, s_{t}^{\text{hard}}) = 4 (1-\epsilon) / M,
\end{align*}
for all $m=M/4+1, ..., M/2$, {\it i.e.,} MDPs from $\mG_{\texttt{ref}}$. Thus, at the last time step, the chance of observing the reward $1$ conditioned on the history that we reached $s_{d}^{\text{hard}}$ with the optimal action sequence $a_{1:d}^*$, is $1/8 + O(\epsilon)$ in hard instances, and $1/8$ in the reference model. Thus, the KL divergence between the two models takes the following form:
\begin{align*}
    &\KL\left( \PP_{\theta_0} (\cdot | \iota_{\text{hard}}, a_{\text{explore}}^*, a_{1:d}^* ), \PP_{\theta} (\cdot | \iota_{\text{hard}}, a_{\text{explore}}^*, a_{1:d}^* ) \right) \\
    &= \sum_{r_d \in \{0,1\}} \PP_{\theta_0} (r_d, s_{1:d}^{\text{hard}}| \iota_{\text{hard}}, a_{\text{explore}}^*, a_{1:d}^*) \cdot \log\left( \frac{\PP_{\theta_0} (r_d, s_{1:d}^{\text{hard}} |\iota_{\text{hard}}, a_{\text{explore}}^*, a_{1:d}^* )}{\PP_{\theta} (r_d, s_{1:d}^{\text{hard}} |\iota_{\text{hard}}, a_{\text{explore}}^*, a_{1:d}^* )} \right) \\
    &= \PP_{\theta_0} (s_{1:d}^{\text{hard}}| \iota_{\text{hard}}, a_{\text{explore}}^*, a_{1:d}^*) \sum_{r_d \in \{0,1\}} \PP_{\theta_0} (r_d | s_{1:d}^{\text{hard}}, \iota_{\text{hard}}, a_{\text{explore}}^*, a_{1:d}^* ) \log\left( \frac{\PP_{\theta_0} (r_d | s_{1:d}^{\text{hard}}, \iota_{\text{hard}}, a_{\text{explore}}^*, a_{1:d}^* )}{\PP_{\theta} (r_d | s_{1:d}^{\text{hard}}, \iota_{\text{hard}}, a_{\text{explore}}^*, a_{1:d}^* )} \right) \\
    &\le \frac{1}{2d} \cdot \KL(\Ber(1/8), \Ber(1/8 + O(\epsilon)) \lesssim \epsilon^2 / M.
\end{align*}

For other inequalities, note that for any trajectory with any $a \neq a^*_{\text{explore}}$, for all $\iota$ and $a_{1:d} \in \mA^{\bigotimes d}$, the marginal distribution is always the same in all hard-instances and the reference model. The marginal distribution is also the same when transitioning to $s^{\text{ref}}_1$ even if $a^*_{\text{explore}}$ is executed at the initial time. Thus, we can focus on the case when the action at the initial time step is $a_{\text{explore}}^*$, and the environment transitions to $s^{\text{hard}}_1$. If this is the case, for all $s_{2:d}$, 
\begin{align*}
    \PP((s_{1}^{\text{hard}}, s_{2:d}), r_d | \iota, a_{\text{explore}}^*, a_{1:d}) &= \sum_{m\in [M/2]} p_m(\iota) \PP((s_{1}^{\text{hard}}, s_{2:d}), r_d | a_{\text{explore}}^*, a_{1:d}, m) \\
    &= \sum_{m\in [M/2]} p_m(\iota) \PP(s_{1}^{\text{hard}} | a_{\text{explore}}^*, m) \PP(s_{2:d}, r_d | s_{1}^{\text{hard}}, a_{1:d}, m).
\end{align*}
Note that 
\begin{align*}
    p_m(\iota) = \frac{\PP(\iota| m)}{\sum_{m'} \PP(\iota| m')},
\end{align*}
and in all models, and since for all $m \in [M/2]$,
\begin{align*}
    \PP(\iota | m) \propto (1 + \alpha \varepsilon_{\iota, m}),
\end{align*}
we can observe that
\begin{align*}
    p_m(\iota) &= \frac{(1+\alpha\varepsilon_{i,m})}{M/2 + \sum_{m'\in[M/2]} (1+\alpha\varepsilon_{\iota,m'}) } = \frac{(1 + O(\alpha))}{M},
\end{align*}
Therefore, in all instances,
\begin{align*}
    \PP (s_{1}^{\text{hard}}, s_{2:d}, r_d | \iota, a_{\text{explore}}^*, a_{1:d}) &= \epsilon \sum_{m\in [M/4]} p_m(\iota) \PP (s_{2:d}, r_d | s_{1}^{\text{hard}}, a_{1:d}, m) \\
    &\quad + (1-\epsilon) \sum_{m\in [M/4+1,M/2]} p_m(\iota) \PP (s_{2:d}, r_d | s_{1}^{\text{hard}}, a_{1:d}, m) \\
    &= \frac{(1+O(\alpha)) \epsilon}{4}  \sum_{m\in [M/4]} \frac{4}{M} \PP (s_{2:d}, r_d | s_{1}^{\text{hard}}, a_{1:d}, m) \\
    &\quad + (1-\epsilon) \sum_{m\in [M/4+1,M/2]} p_m(\iota) \PP (s_{2:d}, r_d | s_{1}^{\text{hard}}, a_{1:d}, m).
\end{align*}
Now comparing this probability between any hard-instance $\theta\in \Theta_{\text{hard}}$ and reference model, note that
\begin{align*}
    &\PP_{\theta_0} (s_{2:d}, r_d | s_{1}^{\text{hard}}, a_{1:d}, m) = \PP_{\theta}(s_{2:d}, r_d | s_{1}^{\text{hard}}, a_{1:d}, m),
\end{align*}
for all $m \in [M/4+1, M/2]$, and 
\begin{align*}
    \sum_{m \in [M/4]} \frac{4}{M} \PP_{\theta_0} (s_{2:d}, r_d | s_{1}^{\text{hard}}, a_{1:d}, m) &= \sum_{m \in [M/4]} \frac{4}{M} \PP_{\theta} (s_{2:d}, r_d | s_{1}^{\text{hard}}, a_{1:d}, m),
\end{align*}
for all $s_{2:d} \neq s_{2:d}^{\text{hard}}$ or $a_{1:d} \neq a_{1:d}^*$, and
\begin{align*}
    &\sum_{m \in [M/4]} \frac{4}{M} \PP_{\theta_0} (s_{2:d}^{\text{hard}}, r_d=1 | s_{1}^{\text{hard}}, a_{1:d}^*, m) = \frac{4}{M} \cdot \frac{1}{8} = \frac{1}{2M}, \\
    &\sum_{m \in [M/4]} \frac{4}{M} \PP_{\theta} (s_{2:d}^{\text{hard}}, r_d | s_{1}^{\text{hard}}, a_{1:d}^*, m) = \frac{4 \cdot \indic{r_d=1}}{M}.
\end{align*}
Therefore, 
\begin{align*}
    &| \PP_{\theta}(s_{1}^{\text{hard}}, s_{2:d}, r_d | \iota, a_{\text{explore}}^*, a_{1:d}) - \PP_{\theta_0}(s_{1}^{\text{hard}}, s_{2:d}, r_d | \iota, a_{\text{explore}}^*, a_{1:d})| \\
    &= O(\alpha\epsilon) \sum_{m \in [M/4]} \frac{4}{M} \PP_{\theta_0} (s_{2:d}, r_d | s_{1}^{\text{hard}}, a_{1:d}, m) + 
    O(\epsilon) \frac{\indic{s_{2:d} = s_{2:d}^{\text{hard}}, a_{1:d}=a_{1:d}^*}}{M},
\end{align*}
and also we note that
\begin{align*}
    &\sum_{m \in [M/4]} \frac{4}{M} \PP_{\theta_0} (s_{2:d}, r_d | s_{1}^{\text{hard}}, a_{1:d}, m) = O(\PP_{\theta_0} (s_{1}^{\text{hard}}, s_{2:d}, r_d | \iota, a^*_{\text{explore}}, a_{1:d})), \\
    &\PP_{\theta_0} (s_{1:d}^{\text{hard}}, r_d | \iota, a^*_{\text{explore}}, a_{1:d}^*) = O(1/M). 
\end{align*}
Therefore, we can ensure that
\begin{align*}
    &| \PP_{\theta}(s_{1}^{\text{hard}}, s_{2:d}, r_d | \iota, a_{\text{explore}}^*, a_{1:d}) - \PP_{\theta_0}(s_{1}^{\text{hard}}, s_{2:d}, r_d | \iota, a_{\text{explore}}^*, a_{1:d})| \\
    &\qquad \le O(\alpha \epsilon) \PP_{\theta_0}(s_{1}^{\text{hard}}, s_{2:d}, r_d | \iota, a_{\text{explore}}^*, a_{1:d}), \\
    &\sum_{s_{2:d}, r_d} | \PP_{\theta}(s_{1}^{\text{hard}}, s_{2:d}, r_d | \iota, a_{\text{explore}}^*, a_{1:d}) - \PP_{\theta_0}(s_{1}^{\text{hard}}, s_{2:d}, r_d | \iota, a_{\text{explore}}^*, a_{1:d})| \le O(\alpha\epsilon),
\end{align*}
for all $a_{1:d} \neq a_{1:d}^*$, and similarly for $a_{1:d}^*$, 
\begin{align*}
    &| \PP_{\theta}(s_{1}^{\text{hard}}, s_{2:d}, r_d | \iota, a_{\text{explore}}^*, a_{1:d}^*) - \PP_{\theta_0}(s_{1}^{\text{hard}}, s_{2:d}, r_d | \iota, a_{\text{explore}}^*, a_{1:d}^*)| \\
    &\qquad \le O(\epsilon) \PP_{\theta_0}(s_{1}^{\text{hard}}, s_{2:d}, r_d | \iota, a_{\text{explore}}^*, a_{1:d}^*), \\
    &\sum_{s_{2:d}, r_d} | \PP_{\theta}(s_{1}^{\text{hard}}, s_{2:d}, r_d | \iota, a_{\text{explore}}^*, a_{1:d}^*) - \PP_{\theta_0}(s_{1}^{\text{hard}}, s_{2:d}, r_d | \iota, a_{\text{explore}}^*, a_{1:d}^*)| \le O(\epsilon).
\end{align*}

Finally, to bound the KL-divergence, using $\log (x) \le x - 1$,
\begin{align*}
    &\KL\left( \PP_{\theta_0} (\cdot | \iota, a_{\text{explore}}^*, a_{1:d} ), \PP_{\theta} (\cdot | \iota, a_{\text{explore}}^*, a_{1:d} ) \right) \\
    &= \sum_{s_{1:d}, r_d} \PP_{\theta_0} (r_d, s_{1:d}| \iota, a_{\text{explore}}^*, a_{1:d}) \cdot \log\left( \frac{\PP_{\theta_0} (r_d, s_{1:d} | \iota, a_{\text{explore}}^*, a_{1:d} )}{\PP_{\theta} (r_d, s_{1:d} | \iota, a_{\text{explore}}^*, a_{1:d} )} \right) \\
    &\le \sum_{s_{1:d}, r_d} \PP_{\theta_0} (r_d, s_{1:d}| \iota, a_{\text{explore}}^*, a_{1:d}) \cdot \left( \frac{\PP_{\theta_0} (r_d, s_{1:d} | \iota, a_{\text{explore}}^*, a_{1:d} )}{\PP_{\theta} (r_d, s_{1:d} | \iota, a_{\text{explore}}^*, a_{1:d} )} - 1 \right) \\
    &= \sum_{s_{1:d}, r_d} \frac{\left|\PP_{\theta_0} (r_d, s_{1:d} | \iota, a_{\text{explore}}^*, a_{1:d} ) - \PP_{\theta} (r_d, s_{1:d} | \iota, a_{\text{explore}}^*, a_{1:d} ) \right|^2}{\PP_{0} (r_d, s_{1:d} | \iota, a_{\text{explore}}^*, a_{1:d} )}  \\
    &= \sum_{s_{2:d}, r_d} \frac{\left|\PP_{\theta_0} (r_d, s_{1}^{\text{hard}}, s_{2:d} | \iota, a_{\text{explore}}^*, a_{1:d} ) - \PP_{\theta} (r_d, s_{1}^{\text{hard}}, s_{2:d} | \iota, a_{\text{explore}}^*, a_{1:d} ) \right|^2}{\PP_{\theta_0} (r_d, s_{1}^{\text{hard}}, s_{2:d} | \iota, a_{\text{explore}}^*, a_{1:d} )},
\end{align*}
which is $O(\alpha\epsilon)^2$ if $a_{1:d} \neq a_{1:d}^*$, and $O(\epsilon^2)$ if $a_{1:d} = a_{1:d}^*$. This concludes the proof of the lemma.

\subsection{Proof of Theorem \ref{theorem:regret_lower_bound}}
\begin{proof}
Suppose any learning strategy (algorithm). Note that an $\epsilon/4$-optimal policy for any given $\theta \in \Theta_{\text{hard}}$ should be able to play the correct action sequence $a_{1:d}^*$ of $\theta$ whenever the prospective  side information is $\iota_{\text{hard}}$. On the other hand,  by pigeon hole principle, for any algorithm $\psi$ with any choice of $K$, there must exist at least one action-sequence $a_{1:d}^*$ and $a_{\text{explore}}^*$ such that, 
\begin{align*}
    \sum_{\iota} \Exs_0[N_{\psi, \iota, a_{1:d}^*}^{\text{explore}} (K)] &= \min_{a \in \mA_{\text{explore}}, a_{1:d}} \left(\sum_{\iota} \Exs_0[N_{\psi, \iota, a_{1:d}}^{a} (K)]\right) \le |\mA_{\text{control}}|^{-(d+1)} \cdot K , \\
    \sum_{\iota \neq \iota_{\text{hard}}, a_{1:d}} \Exs_0[N_{\psi, \iota, a_{1:d}}^{\text{explore}} (K)] &= \min_{a \in \mA_{\text{explore}}} \left( \tssum_{\iota \neq \iota_{\text{hard}}, a_{1:d}} \Exs_0[N_{\psi, \iota, a_{1:d}}^{a} (K)] \right) \le |\mA_{\text{explore}}|^{-1} \cdot K.
\end{align*}
Let $K_0$ be the largest number such that with this choice of $a_{1:d}^*$ and $a^*_{\text{explore}}$, equation \eqref{eq:tv_test_condition} does not hold, i.e.,
\begin{align*}
    \Exs_{\theta_0}[N_{\psi, \iota_{\text{hard}}, a_{1:d}^*}^{\text{explore}} (K_0 + 1)] \gtrsim \frac{1}{\epsilon^2}, \text{ or } \sum_{\iota \neq \iota_{\text{hard}}, a_{1:d}} \Exs_{\theta_0} [N_{\psi, \iota, a_{1:d}}^{\text{explore}} (K_0 + 1)] \gtrsim \frac{1}{\alpha^2 \epsilon^2}. 
\end{align*}
We also note that
\begin{align}
    &N_{\psi, \iota, a_{1:d}}^{a} (K') \ge N_{\psi, \iota, a_{1:d}}^{a} (K), \label{eq:prob1_count_ineq}
\end{align}
for any $K' > K$ with probability 1.

Note that if we play $a \notin \mA_{\text{exploit}}$ whenever $\iota \neq \iota_{\text{hard}}$, we incur at least $(1/8)$-regret. On the other hand, if we do not play $a_{\text{explore}}^*$ or $a_{1:d} \neq a_{1:d}^*$ when $\iota = \iota_{\text{hard}}$, we incur at least $\epsilon/(2M)$-regret. Thus, the total regret of the algorithm in the hard-instance $\theta \in \Theta_{\text{hard}}$ is given by
\begin{align*}
    \text{Regret}_{\theta} (K) &\ge \sum_{a\in\mA, a_{1:d} \neq a_{1:d}^*} \Exs_\theta [N_{\psi, \iota_{\text{hard}}, a_{1:d}}^{a} (K)] \cdot \frac{\epsilon}{2M} + \sum_{\iota \neq \iota_{\text{hard}}, a \in \mA_{\text{explore}}, a_{1:d}} \Exs_\theta [N_{\psi, \iota, a_{1:d}}^{a} (K)] \cdot \frac{1}{8}.
\end{align*}
On the other hand, the regret in the reference model satisfies
\begin{align*}
    \text{Regret}_{0} (K) &\ge \sum_{\iota \neq \iota_{\text{hard}}, a \in \mA_{\text{explore}}, a_{1:d}} \Exs_0 [N_{\psi, \iota, a_{1:d}}^{a} (K)] \cdot \frac{1}{8}.
\end{align*}

Now we consider three cases: 
\paragraph{Case (1).} If $(A/3)^{-d} K_0 \le K \le K_0$, then the condition in equation \eqref{eq:tv_test_condition} cannot be satisfied and therefore, (with proper absolute constants) $\KL \left(\PP^{\psi}_{0} (\tau^{1:K}), \PP^{\psi}_{\text{hard}} (\tau^{1:K}) \right) \le 1/128$, implying $\TV \left(\PP^{\psi}_{0} (\tau^{1:K}), \PP^{\psi}_{\text{hard}} (\tau^{1:K}) \right) \le 1/16$ by Pinsker's inequality. Note that 
\begin{align*}
    \sum_{a\in\mA, a_{1:d} \neq a_{1:d}^*} \Exs_0[N_{\psi, \iota_{\text{hard}}, a_{1:d}}^{a} (K)] = \frac{K}{2} - \Exs_0[N_{\psi, \iota_{\text{hard}}, a_{1:d}^*}^{\text{explore}} (K)] \ge \frac{1}{3} K,  
\end{align*}
and thus, since the sum is always bounded by $K$, with probability at least $1/6$, 
\begin{align*}
    \sum_{a\in\mA, a_{1:d} \neq a_{1:d}^*} N_{\psi, \iota_{\text{hard}}, a_{1:d}}^{a} (K) \ge K/6,
\end{align*}
in the reference model. Here, we used the fact that for any non-negative random variable $A$ that is almost surely bounded by $K$, for all $0 \le x \le K$,
\begin{align*}
    \Exs[A] = \Exs[A| A > x] \cdot \PP(A > x) + \Exs[A| A \le x] \cdot \PP(A \le x) \le K \PP(A>x) + x.
\end{align*}
Therefore in the hard instance $\mM$:
\begin{align*}
    \sum_{a\in\mA, a_{1:d} \neq a_{1:d}^*} N_{\psi, \iota_{\text{hard}}, a_{1:d}}^{a} (K) \ge K/6,
\end{align*}
with probability at least $1/16$, confirming $\sum_{a\in\mA, a_{1:d} \neq a_{1:d}^*} \Exs[N_{\psi, \iota_{\text{hard}}, a_{1:d}}^{a} (K)] \ge K/256$. Thus in this case, 
\begin{align*}
    \text{Regret}_\theta (K) \gtrsim \frac{K\epsilon}{M}. 
\end{align*}

\paragraph{Case (2).}  Suppose $K > K_0$ and $\Exs_0[N_{\psi, \iota_{\text{hard}}, a_{1:d}^*}^{\text{explore}} (K_0 + 1)] \gtrsim \frac{1}{\epsilon^2}$ but $\Exs_0[N_{\psi, \iota_{\text{hard}}, a_{1:d}^*}^{\text{explore}} (K_0)] \lesssim \frac{1}{\epsilon^2}$. Note that by the same argument, we know that
\begin{align*}
    \sum_{a\in\mA, a_{1:d} \neq a_{1:d}^*} \Exs[N_{\psi, \iota_{\text{hard}}, a_{1:d}}^{a} (K_0)] \ge K_0/256,
\end{align*}
and since equation \eqref{eq:prob1_count_ineq} holds with probability 1, we have
\begin{align*}
    \sum_{a\in\mA, a_{1:d} \neq a_{1:d}^*} \Exs[N_{\psi, \iota_{\text{hard}}, a_{1:d}}^{a} (K)] \ge K_0 / 256. 
\end{align*}
On the other hand, to satisfy this condition, we need at least $K_0 \ge \frac{(A/3)^d}{\epsilon^2}$. Thus, plugging this to the regret bound, we have that
\begin{align*}
    \text{Regret}_\theta (K) \ge \frac{K_0}{256} \frac{\epsilon}{2M} \gtrsim \frac{(A/3)^d}{M \epsilon}.
\end{align*}

\paragraph{Case (3).}  Finally, suppose $K > K_0$ and $\Exs_0[N_{\psi, \iota, a_{1:d}}^{\text{explore}} (K_0 + 1)] \gtrsim \frac{1}{\alpha^2 \epsilon^2}$ but $\Exs_0[N_{\psi, \iota, a_{1:d}}^{\text{explore}} (K_0)]  \lesssim \frac{1}{\alpha^2 \epsilon^2}$. Then by construction (due to the choice of $a_{\text{explore}}^*)$, we have 
\begin{align*}
    \sum_{\iota \neq \iota_{\text{hard}}, a \in \mA_{\text{explore}}, a_{1:d}} \Exs_0 [N_{\psi, \iota, a_{1:d}}^{a} (K)] \ge \frac{(A/3)}{\alpha^2 \epsilon^2},
\end{align*}
and thus the regret incurred in the reference model is
\begin{align*}
    \text{Regret}_0 (K) \ge \frac{(A/3)}{\alpha^2 \epsilon^2} \frac{1}{8} \gtrsim \frac{A}{\alpha^2\epsilon^2}.
\end{align*}

Combining the three cases, for any algorithm $\psi$, we can conclude that 
\begin{align*}
    \max_{\theta \in \Theta_{\text{hard}} \cup \{\theta_0\}} \text{Regret}_{\theta} (K) \gtrsim \min \left( \frac{(A/3)^d}{M\epsilon}, \frac{A}{\alpha^2\epsilon^2}, K\epsilon \right).
\end{align*}

\end{proof}

\end{appendices}

\end{document}